%% file: main_paper.tex
\documentclass[10pt,twocolumn,letterpaper]{article}
\usepackage[accsupp]{axessibility}  % Improves PDF readability for those with disabilities.

\usepackage{wacv}
\usepackage{times}
\usepackage{epsfig}
\usepackage{graphicx}
\usepackage{amsmath}
\usepackage{amssymb}
\usepackage{cite}
\usepackage[table]{xcolor}

\usepackage[square,numbers]{natbib}

\usepackage[ruled,vlined]{algorithm2e}

\usepackage{booktabs}

\newcommand{\zhenhua}[1]{\color{red}{zhenhua: }}
\newcommand{\xhdr}[1]{\vspace{4pt}\noindent{\textbf{#1}}}
\newcommand*\samethanks[1][\value{footnote}]{\footnotemark[#1]}
\input{math_commands.tex}

\def\wacvPaperID{****} % Enter the WACV Paper ID here

\wacvfinalcopy % *** Uncomment this line for the final submission

\ifwacvfinal
\fi

\ifwacvfinal
\usepackage[breaklinks=true,bookmarks=false]{hyperref}
\else
\usepackage[pagebackref=true,breaklinks=true,colorlinks,bookmarks=false]{hyperref}
\fi

\begin{document}

%%%%%%%%% TITLE
\title{Semantically Stealthy Adversarial Attacks against Segmentation Models}

\author{Zhenhua Chen\thanks{Equal Contribution} \qquad Chuhua Wang\samethanks[1] \qquad David Crandall \\
Indiana University Bloomington\\
 Bloomington, IN 47405\\
{\tt\small \{chen478, cw234, djcran\}@iu.edu} 
% For a paper whose authors are all at the same institution,
% omit the following lines up until the closing ``}''.
% Additional authors and addresses can be added with ``\and'',
% just like the second author.
% To save space, use either the email address or home page, not both
% \and
% Chuhua Wang\\
% Indiana University Bloomington\\
% 107 S Indiana Ave, Bloomington, IN 47405\\
% {\tt\small cw234@iu.edu}

% \and
% David Crandall\\
% Indiana University Bloomington\\
% 107 S Indiana Ave, Bloomington, IN 47405\\
% {\tt\small djcran@indiana.edu}
}

\maketitle

\ifwacvfinal
\thispagestyle{empty}
\fi

%%%%%%%%% ABSTRACT
\begin{abstract}
Segmentation models have been found to be vulnerable to targeted and non-targeted adversarial attacks. However, the resulting segmentation outputs are often so damaged that it is easy to spot an attack. In this paper, we propose semantically stealthy adversarial attacks which can manipulate targeted labels while preserving non-targeted labels at the same time.  One challenge is making semantically meaningful manipulations across datasets and models. Another challenge is avoiding damaging non-targeted labels. To solve these challenges, we consider each input image as prior knowledge to generate perturbations. We also design a special regularizer to help extract features. To evaluate our model's performance, we design three basic attack types, namely `vanishing into the context,' `embedding fake labels,' and `displacing target objects.' Our experiments show that our stealthy adversarial model can attack segmentation models with a relatively high success rate on Cityscapes, Mapillary, and BDD100K. Our framework shows good empirical generalization across datasets and models.  
\end{abstract}

\begin{figure}[t]
\begin{center}
   \includegraphics[width=1\linewidth]{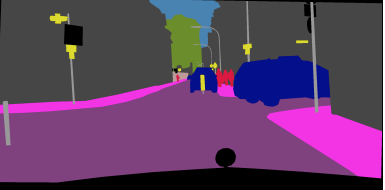}
   \includegraphics[width=1\linewidth]{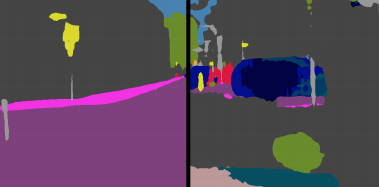}
   \includegraphics[width=1\linewidth]{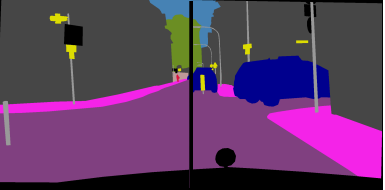}
\end{center}
   \caption{Comparison between stealthy and non-stealthy attacks. Top:  Predictions without attacks. Red represents the `person' label. Middle: Predictions before (left) and after (right) normal attacks in which the incorrect predictions are obvious. Bottom: Predictions before (left) and after (right) semantically stealthy attacks, in which it is difficult to detect an attack since the `person' labels vanished into the environment.}
\label{fig:non-target_attack}
\end{figure}

\begin{figure*}[t]
\begin{center}
   \includegraphics[trim={0cm 6cm 0cm 5cm},clip,width=1\linewidth]{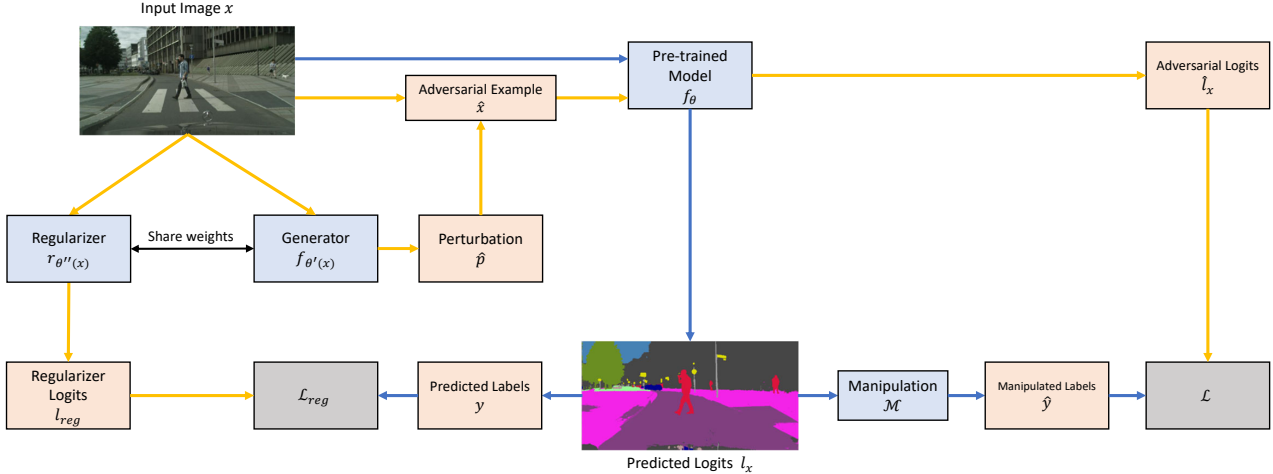}
\end{center}
   \caption{An overview of our attack model. During training, each input image is fed into both generator and regularizer to generate perturbation and regularizer logits. The scaled perturbation is then added to the original image to produce an adversarial example. Pre-trained target model receives both the original and adversarial image, and original predicted logits and adversarial logits are acquired. The predicted logits are manipulated  based on different attack types, and output manipulated labels to calculate adversarial loss. The regularizer logits are paired with predicted labels to calculate regularizer loss.}
\label{fig:workflow}
\end{figure*}

%%%%%%%%% BODY TEXT
\section{Introduction}
It is well-known that neural networks are vulnerable to adversarial attacks.
%perhaps because  networks are roughly linear so that a small perturbation at the input can make a big difference at the output.
For example, we can add imperceptible perturbations to an input image to manipulate the predictions of classifiers~\cite{panda2gibbon, easyfooled, easyfooled1, easyfooled2, adv_trust_region, adv_boosting}, object detectors~\cite{adv_obj_seg}, segmentation models~\cite{adv_seg}, object trackers~\cite{Transferable, IoUAttack, SPARK, EfficientAdv}, edge detectors~\cite{adv_edge}, 3D reconstruction models~\cite{AdvPC, adv_3d}, image caption models~\cite{adv_caption}, face detectors~\cite{adv_face},  embodied agents~\cite{EmbodiedAgents}, video classifiers~\cite{advVideoClassifier}, etc.

Among all the above adversarial attacks, those against segmentation models are relatively difficult because of the thousands of constraints that need to be considered for each input sample. So far, most adversarial segmentation models focus on non-targeted objectives~\cite{adv_obj_seg, adv_seg}, which means that when a segmentation model is attacked, the abnormal predictions are  noticeable. Although there are some targeted adversarial segmentation models, most of them focus on either a static target~\cite{gap, uap_seg} or a particular attribute~\cite{treu2021fashionguided, uap_seg}.

Here we propose semantically ``stealthy'' adversarial attacks against
segmentation models, so that they are not as noticeable. To illustrate this stealthy idea, we design three attack types, 1) making a target object vanish into the context; 2) adding a fake object; 3) replacing a target object with a fake one at a different position. For example, in Figure~\ref{fig:label_vanish}, the `person' labels disappear after being attacked, and at the same time, these pixels are re-classified as `cars.' Since the `car' labels are common in this context, it would be difficult to detect this attack. 

To achieve semantically stealthy adversarial attacks, we face two challenges. On one hand, the targeted labels need to be efficiently relabeled as designed. On the other hand, all other non-targeted pixels' labels should be preserved. Since all the pixels' labels in a
segmentation model have to be considered, the total number of constraints is huge. To tackle this problem we introduce two approaches:
1) an image-dependent perturbation
generation scheme, and 2) a regularizer. Our regularizer is a segmentation model using the same structure as FCN~\cite{fcn}, as Figure~\ref{fig:workflow} shows. The regularizer is introduced based on the following assumption: if a perturbation generator can generate effective noise then it should also be able to predict segmentation labels with high
accuracy since the two tasks need common spatial features. In other
words, the generated features should be task-irrelevant.
%Please
%check the ablation study session for details. 

In addition to the task-related constraints, we also constrain our perturbations to be imperceptible by requiring their infinity norm to be smaller than 10, $\|\hat{p}\|_{\infty} \le 10$). We are also constrained by GPU memory: segmentation models are usually large since they need to predict labels for every pixel, and we have to use another large neural network to fit the complex tasks. Thus, putting both networks during training would consume a large amount of memory. We introduce the concept of `attack efficiency,' namely the ratio of the attack/target model's size.

Our primary contributions are summarized as follows:

\begin{itemize}
  \item We propose an attack framework that can create stealthy, semantically-meaningful attacks against segmentation models. %\djc{what does ``dynamic'' mean in this context? moving?} 
  \item We design three basic types of semantic attacks: removing labels, adding fake labels, and moving labels. Other high-level semantic manipulations can be performed by combining these basic types.  
  \item We evaluate our model's generalization ability across different datasets. 
  \item We propose the concept of parameter-wise efficiency for adversarial models.
\end{itemize}

%-------------------------------------------------------------------------
\section{Related work}
\subsection{Universal adversarial attacks}
The universal adversary might be a solution to overcome the
computationally intensive issue of FGSM-based attacks. Universal
Adversarial Perturbations (UAP) were introduced by~\cite{uap}, and
can generate sample-agnostic perturbations offline and thus can apply attacks in real-time. The key idea of UAP is to find a universal perturbation that can fool all the training/test images. UAP was originally designed for classifiers but can be easily extended to attack segmentation models by, for example, finding a perturbation that maximizes the predictive probability of a target class by iteratively backpropagating the gradients to the input space~\cite{uap_seg}. When the training is finished, the final gradients form the perturbation which is added to each input image during testing time. Another way to obtain UAP is by calculating singular vectors of the Jacobian matrices of the feature maps~\cite{8578991}. It is also possible to use a neural network to generate UAP~\cite{8578182, gap, 8424631}. \citet{Zhang_Benz_Imtiaz_Kweon_2020, Benz_2020_ACCV,
gupta2019method} explore the idea of generating UAPs that
only attack a group of classes and limit the influence
on the remaining classes. Another interesting perspective towards UAP is considering images rather than perturbations as noise~\cite{adv_img_perturb}. As for defending against UAP, adversarial learning is still the most efficient way~\cite{UAP_survey}.

\subsection{Image-dependent adversarial attacks}
Methods based on UAP attack models largely avoid the computationally expensive issue. However, they are not so flexible since each input image, though it can be considered as prior knowledge, is ignored. One straightforward way to solve this issue is by generating perturbations from input images. We can also adopt a GAN-like structure, in which the discriminator corresponds to the target model while the generator is supposed to generate the perturbations. The perturbation generator can start from either random noise or an image. If the perturbation comes from an image, it is called an image-dependent perturbation. For example, \citet{gap} proposes a universal, GAN-like attack model that can attack both segmentation models and classifiers.

\subsection{Semantic adversarial attacks}
Semantic/targeted adversarial attacks are related to our work. \citet{ColorFool} propose a content-based black-box adversarial attack through manipulating background colors. \citet{SemanticAdvAttacks} creates semantic adversarial
attacks against classifiers by manipulating specific attributes.
These types of attacks are somewhat stealthy but their target
models are classifiers. \citet{treu2021fashionguided} proposed a method to overlay generated adversarial texture on the clothing  of a person to fool segmentation networks. \citet{uap_seg} is most alike our work in which an adversarial attack for hiding person labels is proposed. However, \citet{uap_seg} adopt universal perturbations and tested only on one dataset.

\subsection{Physical adversarial attacks}
Apart from adding perturbations, it is also possible to add physical attributes to a scene into order to attack. For example, \citet{glass_fool_face} can fool a face recognizer by adding glasses to a face, while \citet{semanticadv} manipulates face attributes (like hair color, etc.). \citet{advCamouflage} manipulate classifiers' predictions by camouflaging physical adversarial examples into natural styles. These methods used to generate physical adversarial examples are weakly related to our work.

\section{Our Approach}
Many studies~\cite{NIPS2017_d494020f,adv_obj_seg,xiao2018characterizing} show that complex tasks such as semantic segmentation are vulnerable to adversarial attacks, and that generated subtle perturbations to inputs can completely break the prediction outputs.   We present a stealthy adversarial attack approach that not only can hide an attack behavior, but also keeps other non-targeted labels correctly classified. We realize our goal by designing a regularized pipeline to generate image-dependent perturbations, as shown in Figure~\ref{fig:workflow}.

In this section, we first give an overview of the adversarial attack problem and our proposed stealthy approach. We then introduce how the generator and regularizer work.

\subsection{Problem definition}
Let $f_{\theta}$ be the target segmentation model trained on some dataset 
with (image, label) pairs $\displaystyle \gD = \{(x, y)\}$ where $x \in \displaystyle
\R^{h\times w \times c}$, $y \in \displaystyle \R^{h\times w}$ and $h$, $w$, and $c$ represent
the height, width, and number of image channels, respectively. The perturbation $\hat{p} \in \displaystyle \R^{h\times w \times c}$ has the same size as the input image that can be acquired by the generator $f_{\theta^{'}(x)}$. An adversarial example $\hat{x} \in \displaystyle \R^{h\times w \times c}$ is the addition of the perturbation $\hat{p}$ and the input image $x$. We need to map both $x$ and $\hat{x}$ to logits through $f_{\theta}$ during training, 

\begin{equation} \label{eq:model}
l_x =  f_{\theta}(x), \; \hat{l_x} =  f_{\theta}(\hat{x}).
\end{equation}

% \begin{equation} \label{eq:model}
% \hat{l_x} =  f_{\theta}(\hat{x}).
% \end{equation}

Apart from $l_x$ and $\hat{l_x}$, we also train a regularizer that is supposed to predict segmentation labels,

\begin{equation} \label{eq:model}
l_{reg} =  r_{\theta^{''}}(x).
\end{equation}
\noindent
Finally, we pair $l_x, \hat{l_x}, l_{reg}$ with their corresponding labels $y, \hat{y}, y$ to calculate the loss. 
% such that, 1) for a target class $c$, all the adversarial labels of $c$ 
% (\ie $y^{tgt}_{c}$), are different from the predicted labels of $c$ (\ie $\hat{y}_{c}$); 2) for each pixel of $y^{tgt}_{c}$, it becomse the class with the second highest probability of that pixel; 3) the rest of the adversarial labels $y^{tgt}- y^{tgt}_{c}$ remains the same as in $\hat{y} - \hat{y}_{c}$.

% \begin{figure}[!]
% \begin{center}
%   \includegraphics[trim={3cm 3cm 3cm 1cm},clip,width=1\linewidth]{images/structure.pdf}
% \end{center}
%   \caption{The detailed structure of the generator. 3, 64, 128, 256 are number of channels.}
% \label{fig:structure}
% \end{figure}

\subsection{Model overview}
As Figure~\ref{fig:workflow} shows, our framework contains a target model, a generator, and a regularizer. Input images $x$ are fed into the generator $f_{\theta^{'}}$ to produce perturbations. The regularizer (which shares the same backbone $f_{\theta^{'}}$ with the generator) is trained to learn the segmentation task. Both original input images and the corresponding perturbated images are paired and fed to the target model $f_{\theta}$ to generate logits $l_x, \hat{l_x}$. After we map $\hat{l_x}$ to $\hat{y}$, $l_x$ and $\hat{y}$ are fed into the loss function for training.

During training, we first load the pre-trained target model $\displaystyle f_{\theta}$ and freeze it to prevent the weights from updating during
back-propagation. Then the generator takes an image sample and outputs a perturbation image. The regularizer generates logits to fit the original segmentation task. The adversarial image is obtained by adding the scaled perturbation to the original image. Both the original and adversarial are fed into $\displaystyle f_{\theta}$ and output the original and the adversarial logits respectively. Depending on attack designs, we pair manipulated labels with the predicted labels to calculate the adversarial loss. At the same time, we pair the logits generated by the regularizer with the ground truth segmentation labels to calculate the regularizer loss. Finally, both cross-entropy losses from the regularizer and the generator are backpropagated to the generator space. The whole algorithm is summarized in Algorithm~\ref{algorithm}. 

\begin{algorithm*}\label{algorithm}
$\mathbf{Input:}$ input image $x$, Target model $f_{\theta}$ with parameters $\theta$. Generator $f_{\theta^{'}}$ with parameters $\theta^{'}$, regularizer $r_{\theta^{''}}$ with parameters $\theta^{''}$, Stealthy label mapper $\gM$.

% Regularizer weighting factor $\lambda_{0}$, Weighting factor for targeted pixels $\lambda$, Logits of the regularizer $l_{gen}$. \\

$\mathbf{Output:}$ Adversarial perturbations $\hat{p}$, Adversarial input image $\hat{x}$ \\
 Initialize the generator $f_{\theta^{'}}$, Initialize the regularizer $r_{\theta^{''}}$  \\
 \While{not converge}{
  Generate regularizer logits $l_{reg}=r_{\theta^{''}}(x)$, and perturbations $\hat{p}=f_{\theta^{'}}(x)$ \;
  Obtain an adversarial image $\hat{x} = x + \hat{p}$\;
  Generate the clean logits of the target model $l_x = f_{\theta}(x)$\;
  Generate the dirty logits of target model $\hat{l_x} = f_{\theta}(\hat{x})$\;
  Create semantically stealthy labels $\hat{y} = \gM(l_x)$ \;
  Calculate the regularizer loss $\gL_{reg}(y, r_{\theta^{''}}(x))$\;
  Calculate the adversarial loss $\displaystyle \gL(l_x, \hat{l_x})$\;
  Back-propagate the gradients\;
 }
 \caption{Semantically Stealthy Adversarial Attack.}
\end{algorithm*}

\subsubsection{Generator \& regularizer}
We adopt two versions of generators to deal with different sizes of segmentation models: a Unet~\cite{ronneberger2015unet} network (as shown in Figure~\ref{fig:structure}), and a FCN network structure~\cite{fcn} with ResNet-101 backbone~\cite{resnet}. The generator $f_{\theta^{'}}$ is to map an input image to perturbations, as shown in Figure~\ref{fig:structure}. At the same time, we have a regularizer $r_{\theta^{''}}$ which is responsible for predicting the semantic
segmentation labels of the original image. The potential assumption is that if a generator can generate perturbations that serve a particular target, then the embedded features that come from the generator should also have a good knowledge of the spatial relationship within each input image. In other words, the two branches of the generator (one for semantic segmentation label, the other one for perturbations) are complementary with each other. After the perturbation is acquired, we forward both the original image and the adversarial image through the target model to obtain predicted and adversarial labels. We use a cross-entropy loss to regulate the generator and help to preserve labels of non-targeted pixels. The adversarial labels are manipulated (see Section~\ref{attacking_type}) and used to calculate the adversarial loss (see \ref{adv_softmax} for details).

\subsubsection{Attack types} \label{attacking_type}
In general, we achieve stealthiness via manipulating targeted pixels' labels as designed and at the same time preserving untargeted pixels' labels. During attack, we map `clean' labels to `semantically stealthy' labels, 

\begin{equation} 
\hat{y} = \gM (l_x) 
\end{equation}

We design three attack types to manipulate the predicted labels and validate our idea. We believe that these three attack types are basic and combining these attack types can be used to generate other higher-level semantically attack types: Type \#1, remove the target class, Type \#2, generate a class label from an external source, and Type \#3, combine the previous two types. More specifically, 

\begin{figure}[t]
\begin{center}
   \includegraphics[width=1\linewidth,height=0.5\linewidth]{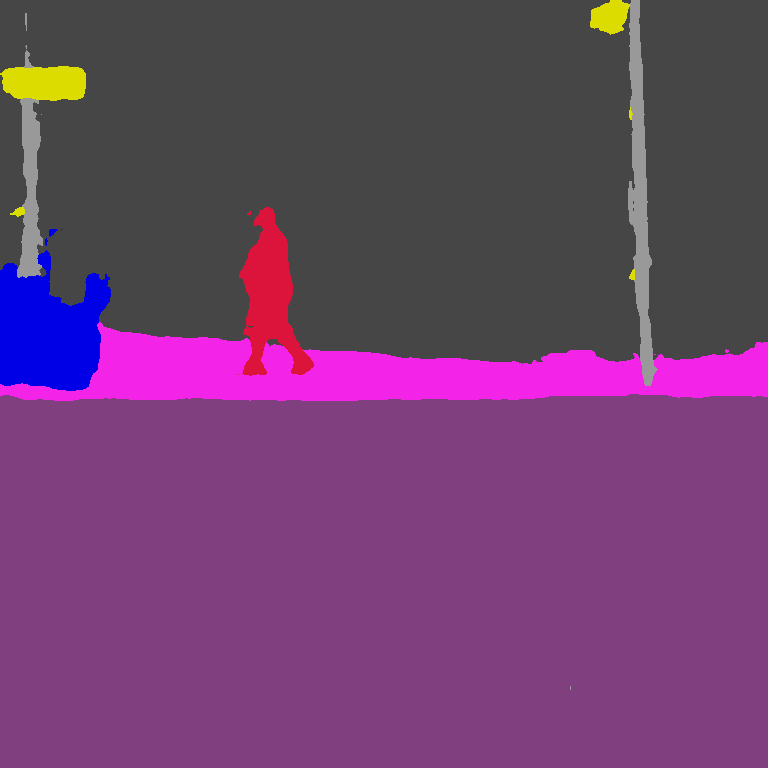}
   \includegraphics[width=1\linewidth,height=0.5\linewidth]{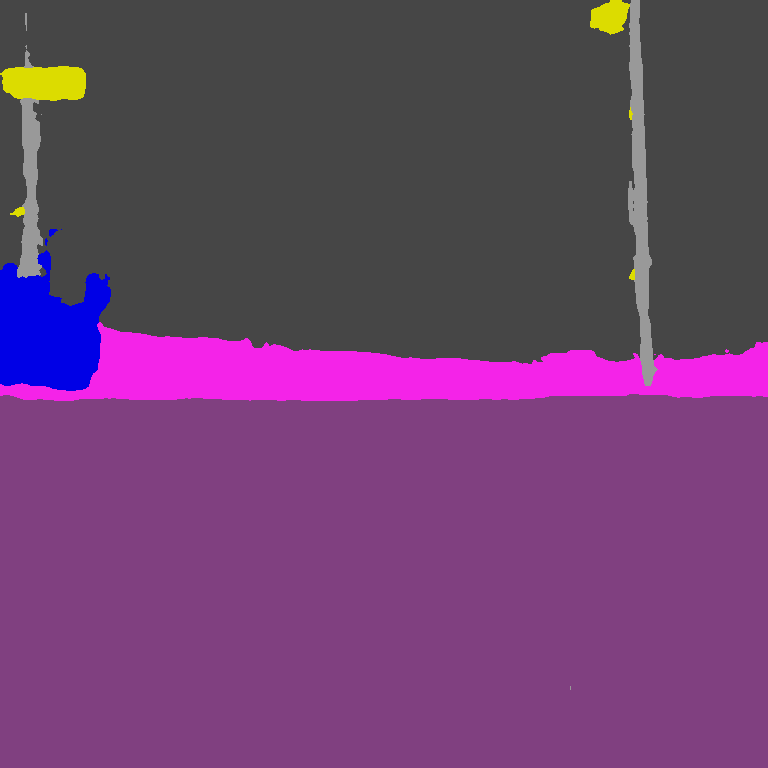}
\end{center}
   \caption{How we convert the original predictions to the ``adversarial ground truth." The `person' labels (the red area) near the wall are converted to label `wall' while those near the sidewalk are converted to label `sidewalk.'}
\label{fig:adv_groundtruth}
\end{figure}

\xhdr{Type \#1:} The target model is made to fail to predict a designated class. As a result, the original label has disappeared or blended into the surrounding context. The most straightforward way of achieving this goal is to assign each target pixel its neighbor's label by clustering. However, clustering is very time-consuming. We thus adopt a heuristic approach by replacing each target pixel's original label with its second likely label in the original prediction. %\djc{not sure what ``second possible label'' means?}
Practically, it works well, as Figure~\ref{fig:adv_groundtruth} shows.  

\xhdr{Type \#2:} The labels of a set of referred pixels are transformed into a specific class.
Visually, this makes it appear like an attacker creates a set of new 
labels that comes from nowhere. We manipulate the predicted label by adding 
a new mask from an external source.

\xhdr{Type \#3:} This task can be achieved by combining the attacks of  Type \#1 and Type \#2: removing the target class labels in one position and then create fakes class labels in another position. 

\subsubsection{Loss}\label{adv_softmax}
We design our loss function to make sure that the target label is manipulated from $y$ to $\hat{y}$, as Equation~\ref{eq:loss} shows. $T_k$ equals  $1$ if the pixel $k$ is one of our target pixels, and otherwise is $0$. Variables $n$, $c'$, $m$ are the batch size, channel size (total number of categories), and the total number of pixels in each image, while $p_{i, j, k}$ represents the probability of each pixel belonging to $j$ which comes from either $l_{x} \in \displaystyle \R^{m\times c'}$ (the logits forms of $y$) or $\hat{l_x} \in \displaystyle \R^{m\times c'}$. The loss function $\mathcal{L(\cdot)}$ is shown in Equation~\ref{eq:loss}.

\begin{equation} \label{eq:loss}
\begin{split}
\mathcal{L}\left(\hat{l_x}, \hat{y}\right) = \min \sum_{i=0}^{n}\sum_{k=0}^{m} T_k (-\log p_{i, \hat{y}_k, k} \\  + \left(1 - T_k\right) \left( -\log \left( p_{i, \hat{y}_k, k} \right) \right).
\end{split}
\end{equation}
Assuming the logits that come from the generator is $l_{reg}$, then the regularizer loss can be summarized as, 
\begin{equation} \label{eq:regularizer}
\mathcal{L}_{reg}(l_{reg}, y) = \min \sum_{i=0}^{n}\sum_{k=0}^{m} -\log p_{i, y_k, k}.
\end{equation}
We introduce $\lambda_0$ to be the weight of regularizer loss. The total loss term is,
\begin{equation} \label{eq:total_loss}
\begin{split}
\mathcal{L}_{total} = \mathcal{L} + \lambda_{0} * \mathcal{L}_{reg}.
\end{split}
\end{equation}
It is common practice to limit the infinity norm of the perturbations to achieve imperceptibility, as shown in~\cite{panda2gibbon}. If not specified, $\xi$ is always equal to 10,

\begin{equation} \label{eq:perturbation}
\|\hat{p}\|_{\infty} \le \xi.
\end{equation}

% \begin{table*}[t]
% \centering
% \begin{tabular}{lccccccc}
%  \toprule
%  & \multicolumn{3}{c}{manipulated (\%)}&  & \multicolumn{3}{c}{preserved (\%)}\\
%  \cmidrule{2-4}  \cmidrule{6-8} 
%  Model & Type1 & Type2 & Type3 &&  Type1 & Type2 & Type3 \\  
% \midrule
%  Cityscapes  &  74.71\% &  91.27\%  & 67.73\%  &&   91.40\% & 92.76\% & 88.74\%  \\
%  %83.01\% 
%  BDD100K  &  66.56\% &  68.69\%  & 66.54\%  &&   97.09\%& 92.46\% & 90.58\% \\
%  %& 80.32\%\\
%  Mapillary  &  68.39\% &  78.67\%  &  56.07\% &&  96.41\% & 88.62\%& 85.42\%\\
%  %& 76.46\%\\
%  \bottomrule
% \end{tabular}
%  \caption{The performance of our framework on three street-view datasets. Each item represents a success rate for type `manipulated' or `preserved' in one of the three datasets.} \label{tab:fcn_result}
% \end{table*}

\begin{table*}[t]
\centering
\begin{tabular}{lccc}
 \toprule 
 Dataset & Type\#1 & Type\#2 & Type\#3   \\ 
\midrule
 Cityscapes  &  \textcolor{red}{74.71\% / 91.40\%}  &  91.27\% / 92.76\%  & 67.73\% / 88.74\%  \\
 
 BDD100K  &  66.56\% / 97.09\% &  68.69\% / 92.46\%  & 66.54\% / 90.58\%   \\
 
 Mapillary  &  68.39\% / 96.41\% &  78.67\% / 88.62\% &  56.07\% / 85.42\%  \\
 \bottomrule
\end{tabular}
 \caption{The performance of our framework on three street-view datasets. Each item represents success rates for type `manipulated' / `preserved' in one of the three datasets. For example, the top-left item (red-colored) means that the attack model trained on Cityscapes can make \textcolor{red}{74.71\%} targeted pixels vanish into the background while making \textcolor{red}{91.40\%} non-targeted pixels preserve their original labels. All the target models are trained by FCN.} \label{tab:fcn_result}
\end{table*}

\begin{figure*}[t]
\begin{center}
    \includegraphics[width=0.3\linewidth]{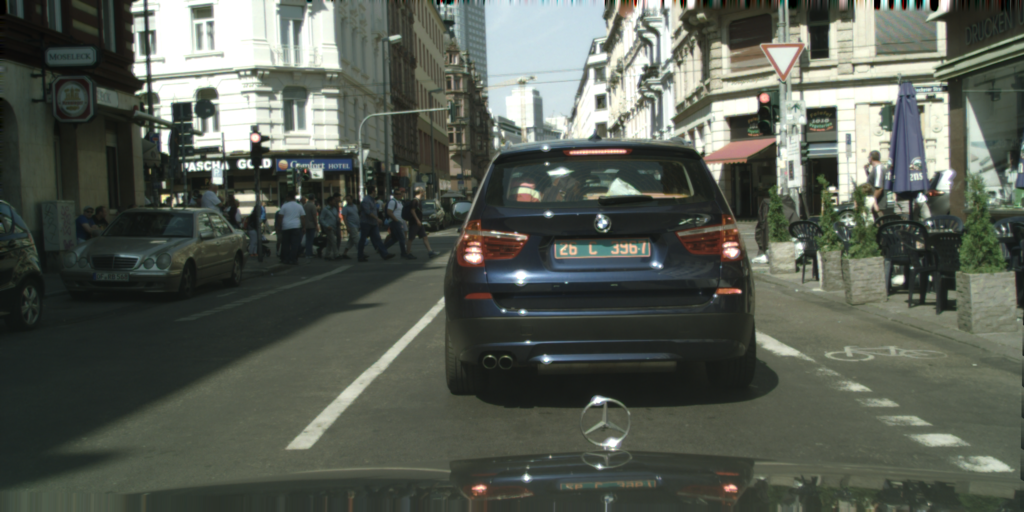}
    \includegraphics[width=0.3\linewidth]{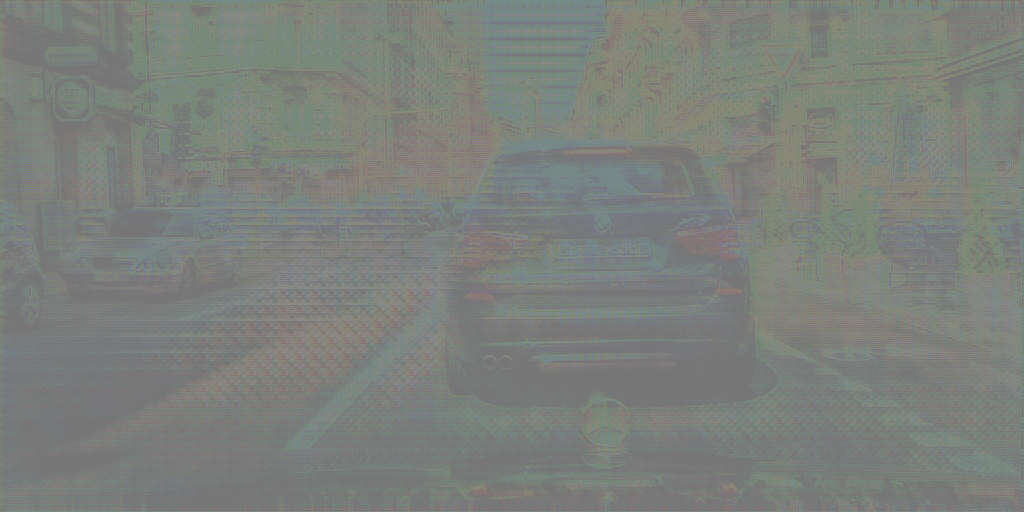}
    \includegraphics[width=0.3\linewidth]{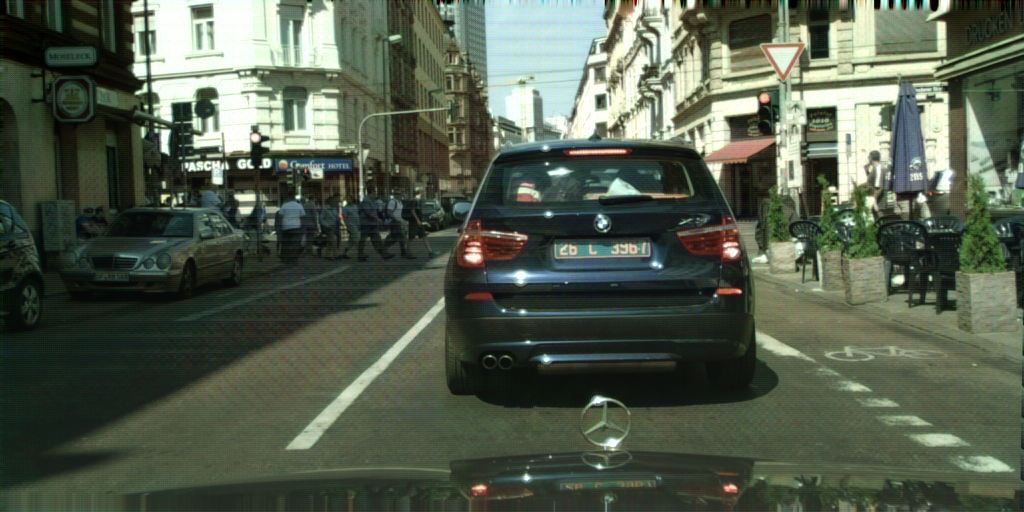}
    \includegraphics[width=0.3\linewidth]{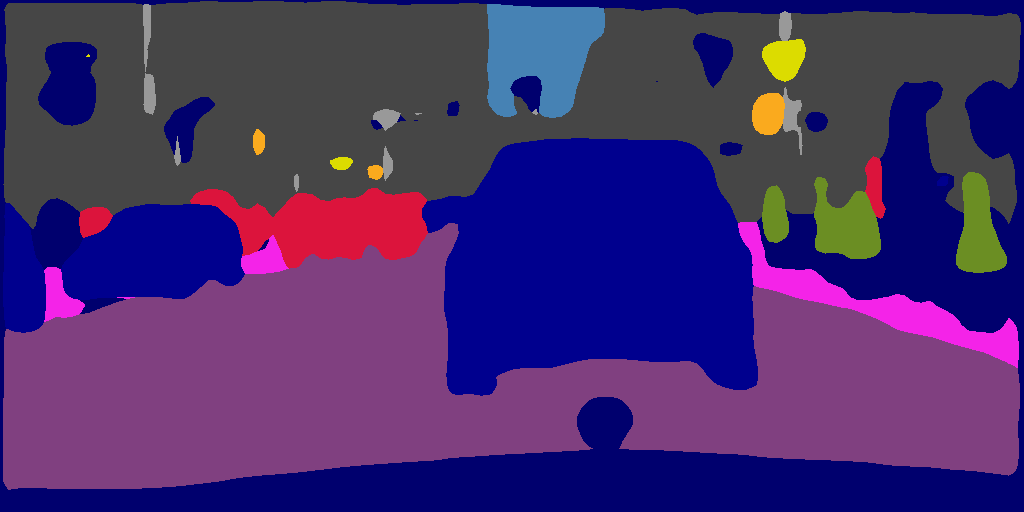}
    \includegraphics[width=0.3\linewidth]{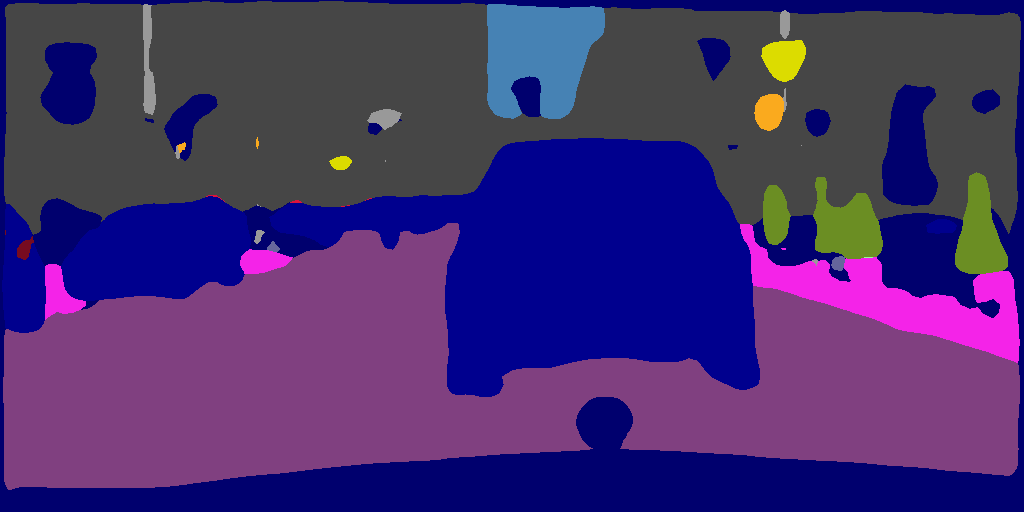}
\end{center}
   \caption{Type \#1 attack on Cityscapes. The `person' labels vanish into the background. Top left: The input image. Top middle: The perturbations. Top right: The input image + perturbations. Lower left: Normal predictions. Lower right: Predictions after attacks.}
\label{fig:label_vanish}
\end{figure*}

% \begin{table*}[t]
% \centering
% \begin{tabular}{lccc}
%  \toprule 
%  Models & Type\#1 & Type\#2 & Type\#3   \\ 
% \midrule
%  DLV3+-MobileNets~\cite{deep_lab_v3_plus}  &  56.14\% / 88.49\% &  62.68\% / 83.98\%  & 35.34\% / \%   \\
 
%  DLV3+-ResNet101~\cite{deep_lab_v3_plus}  &  47.08\% / 93.26\% &  \% / 99.18\% &  \% / \%  \\
%  \bottomrule
% \end{tabular}
%  \caption{The performance of our framework on two segmentation models. Each item represents success rates for type `manipulated' / `preserved' in one of the two models. Both models are trained on CityScapes.} \label{tab:cross_model}
% \end{table*}

\section{Experiments}

\subsection{Datasets \& Metrics}
We choose three street-view datasets as the testing set since all our attack types focus on pedestrians. Specifically, we use Cityscapes
(2975 training images, 1525 testing images)~\cite{cityscapes},  
Mapillary Vistas Dataset (18K training images, 5K testing
images)~\cite{mapillary}, and BDD100K (7K training image, 2K testing images)~\cite{bdd100k}. 

We evaluate our model by using \textit{success rate} as the performance metric. Each success rate is divided into two parts: manipulated and preserved rate. Manipulated rate is defined as the percentage of labels that are manipulated successfully for targeted pixels while the preserved rate is defined as the percentage of labels that are preserved 
for non-targeted pixels.

\begin{figure*}[t]
\begin{center}
    \includegraphics[width=0.3\linewidth]{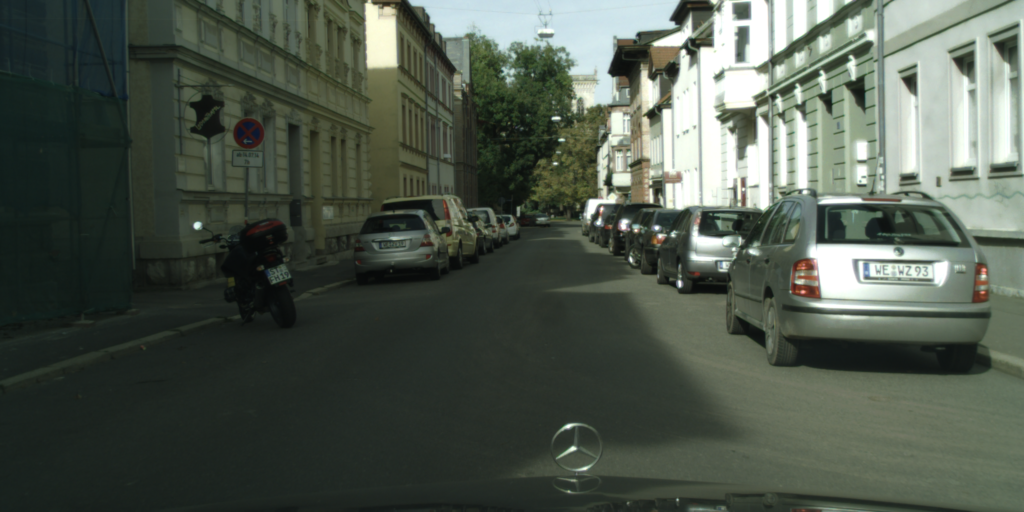}
    \includegraphics[width=0.3\linewidth]{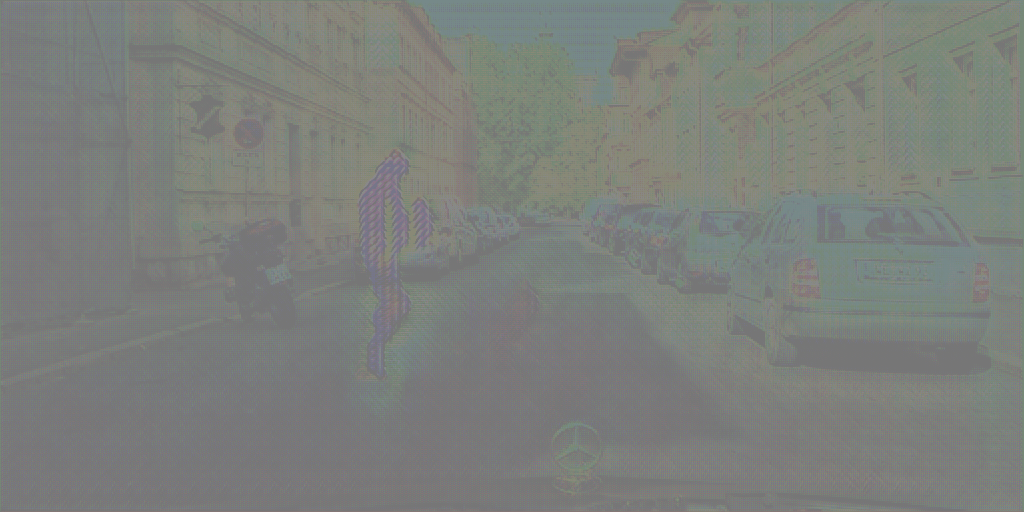}
    \includegraphics[width=0.3\linewidth]{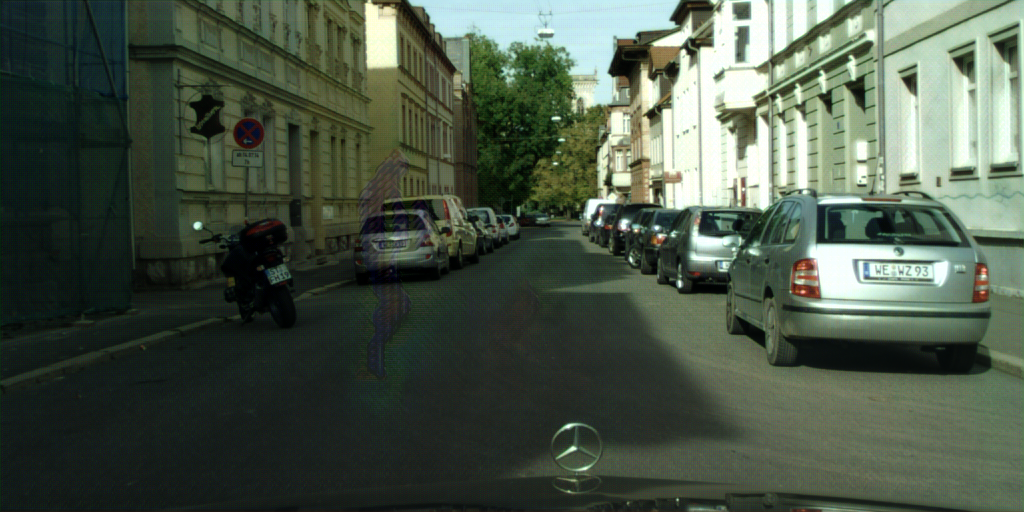}
    \includegraphics[width=0.3\linewidth]{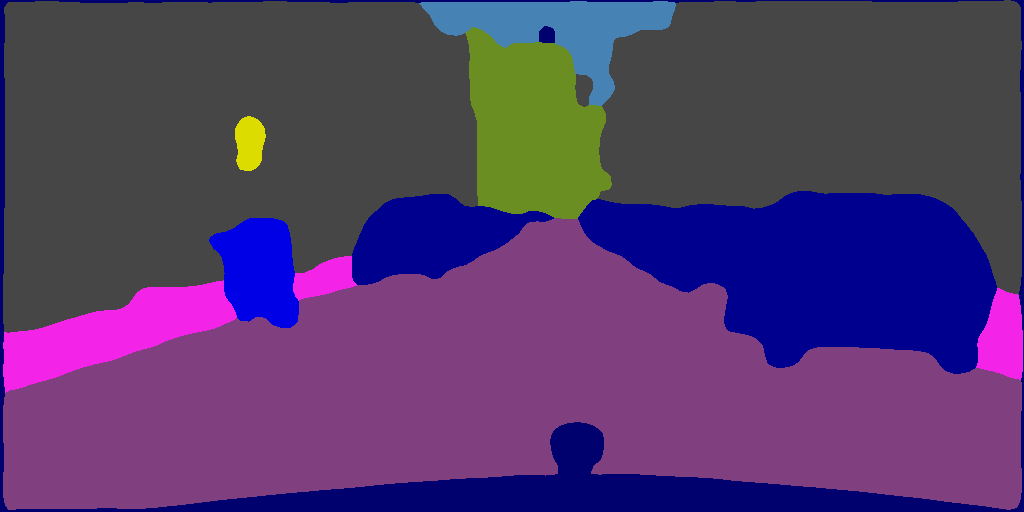}
    \includegraphics[width=0.3\linewidth]{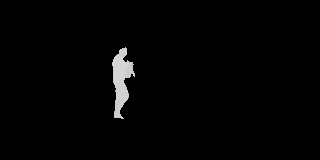}
    \includegraphics[width=0.3\linewidth]{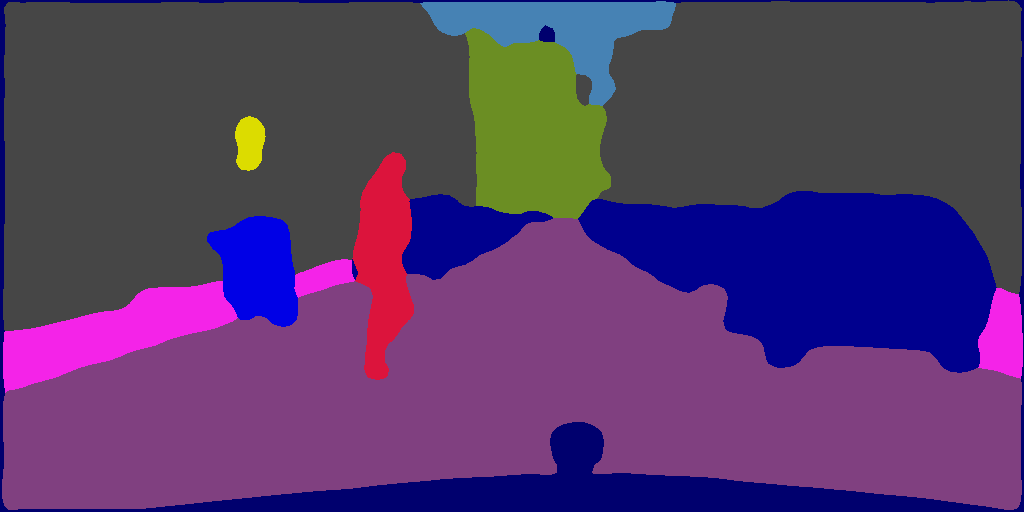}
    
\end{center}
   \caption{Type \#2 attack on Cityscapes. Fake `person' labels are embedded into the prediction domain. Top left: The input image. Top middle: The perturbations. Top right: The input image + perturbations. Lower left: Normal predictions. Lower middle: The fake label mask. Lower right: Predictions after attacks. }
\label{fig:fake_label}
\end{figure*}

\subsection{Implementation details}
We implement our network in PyTorch~\cite{pytorch} and perform all
experiments with Nvidia Titan Xp Pascal GPUs. In Section~\ref{gen_across_datasets}, we use FCN-8s as the target segmentation model, and pre-train the model on all three datasets. We resize each input image to 512$\times$1024 during training. For Section~\ref{en_across_models}, we use CityScapes as our train/test dataset. Each input image is loaded at its original size and then cropped randomly to 768$\times$768 during training. For all the models, we choose \textit{Person} and \textit{Rider} as our target labels. The weight of our regularizer $\lambda_0$ is set to 1e-2. The learning rate is 1e-4, and batch size is set to 16.

\subsection{Quantitative analysis}
Table~\ref{tab:fcn_result} shows the manipulated/preserved rate of attacking the target model~\cite{fcn}. We observe that manipulating non-targeted labels into target labels (Type \#2) seems to be easier than blending target labels into non-targeted labels (Type \#1). We conjecture that the model does not need to consider the context information in the former task, thus leading to a higher manipulated rate. However,
our results show that preserving non-targeted labels in attack type \#2 
is more difficult than in attack type \#1. 
We speculate that this might be because  it is challenging
to preserve surrounding information when the manipulated pixels
do not fit in  context. The results also suggest that combining the two
tasks (type \#3) decreases the overall performance.

\begin{figure*}[t]
\begin{center}
    \includegraphics[width=0.3\linewidth]{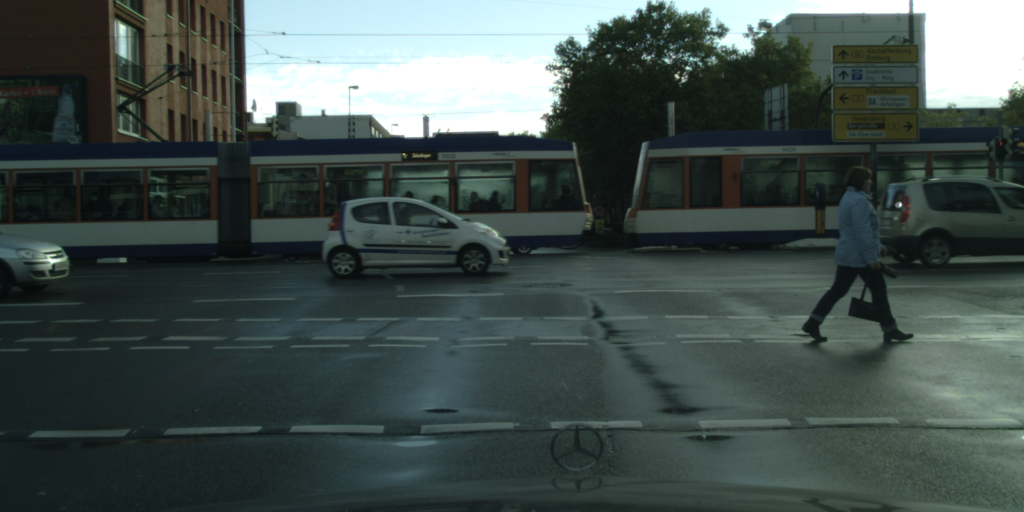}
    \includegraphics[width=0.3\linewidth]{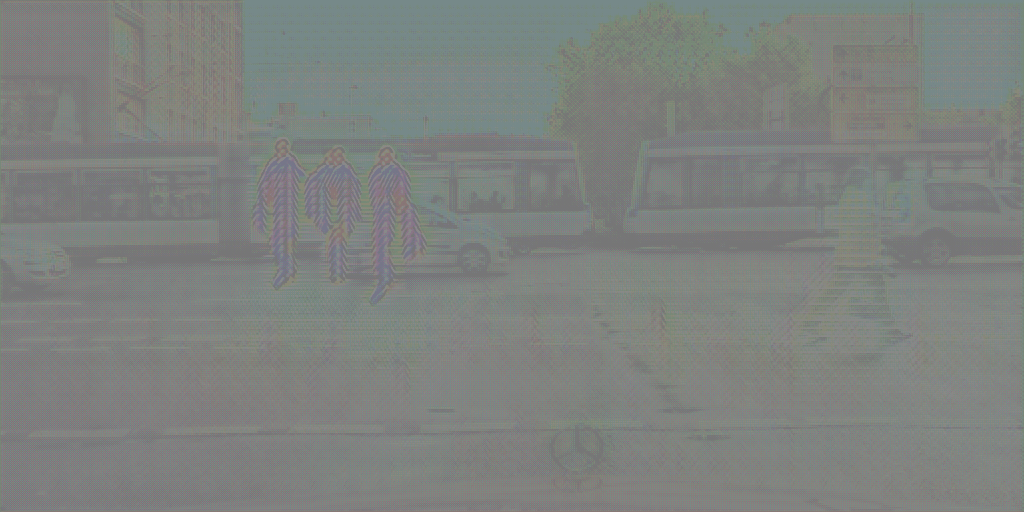}
    \includegraphics[width=0.3\linewidth]{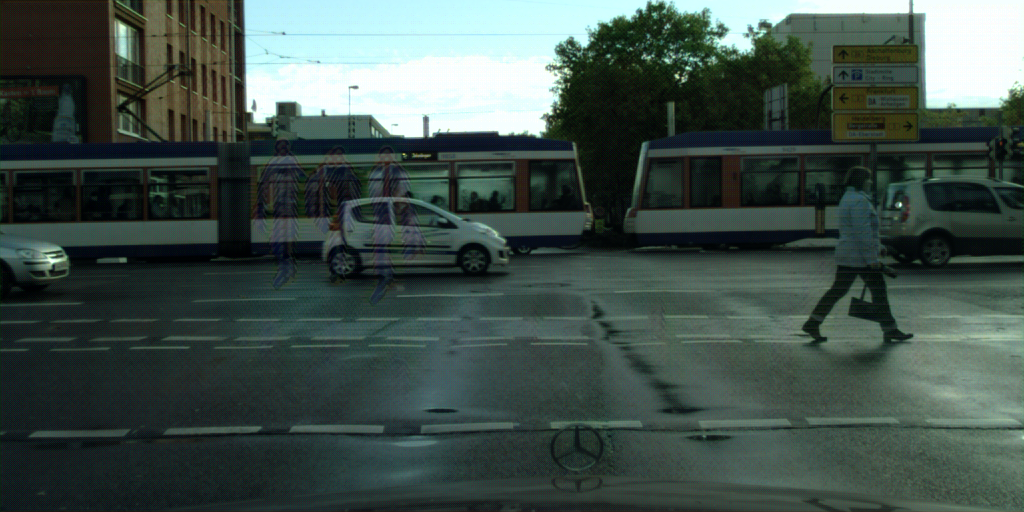}
    \includegraphics[width=0.3\linewidth]{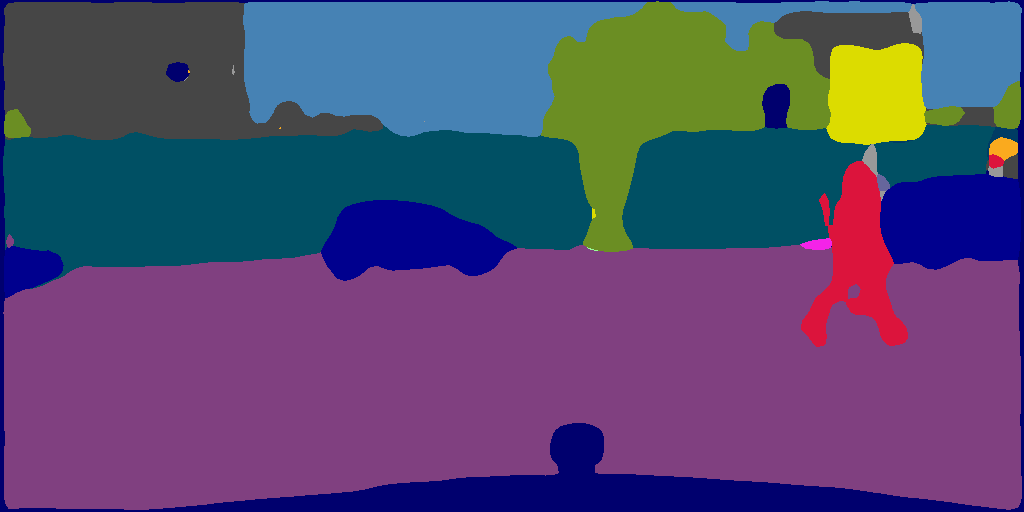}
    \includegraphics[width=0.3\linewidth]{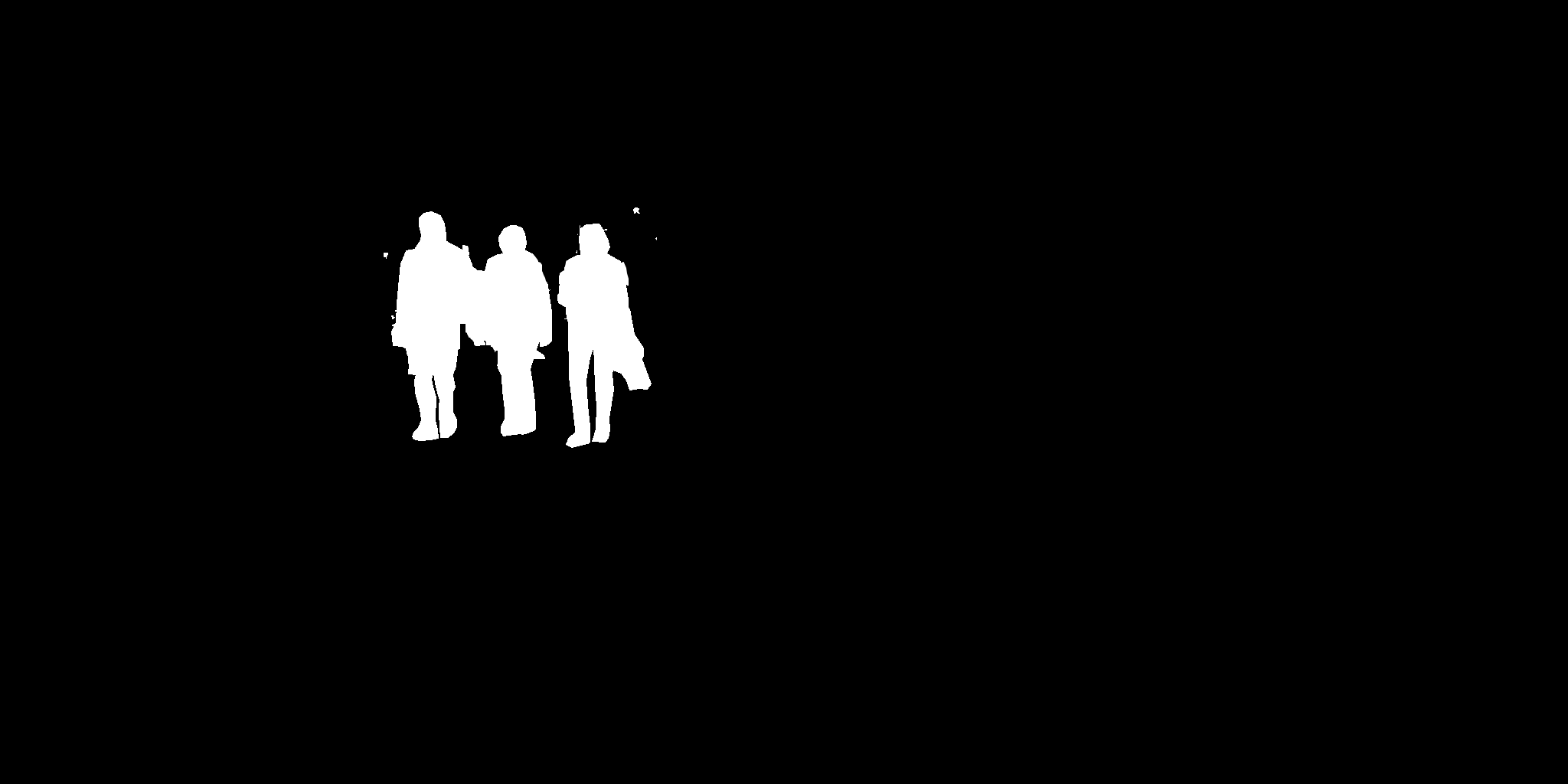}
    \includegraphics[width=0.3\linewidth]{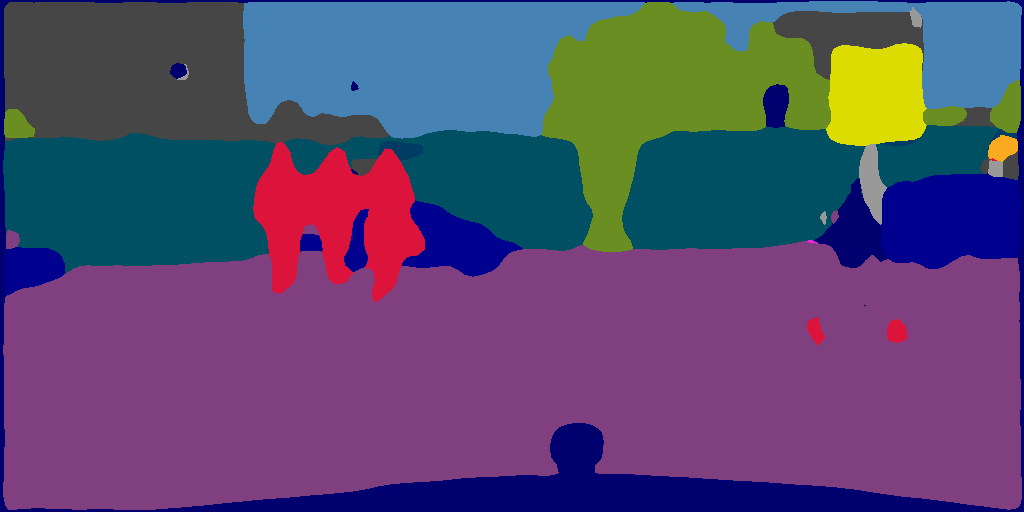}
    
\end{center}
   \caption{Type \#3 attack on Cityscapes. The target `person' labels disappear and the fake `person' labels are embedded. Top left: The input image. Top middle: The perturbations. Top right: The input image + perturbations. Lower left: Normal predictions. Lower middle: The fake label mask. Lower right: Predictions after attacks.}
\label{fig:replace_label}
\end{figure*}

\subsection{Qualitative analysis}
Figure~\ref{fig:label_vanish} shows the results of Type \#1 attack. 
The person labels are converted to building, car, sidewalk, and road labels. There are some `residuals' left in the original position. Figure~\ref{fig:fake_label} shows
the results of Type \#2 attack. We use the binary mask of person label to train our generator. The perturbation includes highlights around the boundary of added person labels, and the target model classifies these non-targeted pixels as person labels. Figure~\ref{fig:replace_label} shows the results of combining the previous two types of attacks. The person labels in the original prediction are not completely removed and three fake `person' regions are embedded. Comparing to the above two tasks, Type \#3 is the most difficult among them.

\begin{table}[!]
\centering
\begin{tabular}{lcccc}
 \toprule
 Datasets & Cityscapes & BDD100K &  Mapillary\\  
\midrule
Cityscapes  &  96.34\% & \textcolor{red}{95.44 \%}  & 92.96\%\\
BDD100K &  97.24\% &  97.86\%  & 96.56\% \\
Mapillary &  96.86\% &  96.09\%  & 97.41\% \\
 \bottomrule
\end{tabular}
 \caption{Our model's generalization across datasets on Type\#1 attack. Each item is the success rate within/across datasets. For example, the red-colored \textcolor{red}{95.44\%} indicates that the attack model trained on Cityscapes can achieve a success rate of \textcolor{red}{95.44\%} against a segmentation model trained on BDD100K.}\label{generalization}
\end{table}

\begin{table}[!]
\footnotesize
\centering
\begin{tabular}{lccc}
 \toprule
Models & DLV3+-MobileNets & DLV3+-ResNet101 \\  
\midrule
Type \#1 &&\\
DLV3+-MobileNets  &  84.36\% &  \textcolor{red}{89.16\%}  \\
DLV3+-ResNet101 &  85.49\% &  89.92\%   \\
\midrule
Type \#2 &&\\
DLV3+-MobileNets  &  87.42\% &  90.67\%   \\
DLV3+-ResNet101 &  96.63\% &  96.80\%   \\
\midrule
Type \#3 &&\\
DLV3+-MobileNets  &  80.91\% &  83.35\%  \\
DLV3+-ResNet101 &  82.82\% &  84.80\%   \\
 \bottomrule
\end{tabular}
 \caption{Our model's generalization across attack models trained on CityScapes. Each item is the overall success within/across models. For example, for type \#1 attack, the red-colored \textcolor{red}{89.16\%} indicates that the attack model trained on DeepLabV3Plus-MobileNets can achieve a success rate of \textcolor{red}{89.16\%} against a target model pretrained on DeepLabV3Plus-ResNet101.}\label{gen_across_models}
\end{table}

% \begin{table}[!]
% \centering
% \begin{tabular}{lccc}
%  \toprule
%  Models & DeepLabPlusV3-MobileNet & DeepLabPlusV3-ResNet101\\  
% \midrule
% DeepLabPlusV3-MobileNet  &  96.34\% & \textcolor{red}{95.44 \%}  \\
% DeepLabPlusV3-ResNet101 &  97.24\% &  97.86\%   \\
%  \bottomrule
% \end{tabular}
%  \caption{Our model's generalization across datasets on Type\#1 attack. Each item is the success rate within/across datasets. For example, the red-colored \textcolor{red}{95.44\%} indicates that the attack model trained on Cityscapes can achieve a success rate of \textcolor{red}{95.44\%} against a segmentation model trained on BDD100K.}\label{generalization}
% \end{table}

\subsection{Generalization across datasets} \label{gen_across_datasets}
It is desirable to check the generalization ability of our model
across different datasets. In other words, we want to find out whether our model can achieve good performance
on unseen distributions. Specifically, we train an adversarial model (model \#1 in Table~\ref{tab:param_efficiency}) on one dataset (e.g. Cityscapes) to attack the segmentation models trained on two other datasets (e.g. Mapillary, and BDD100K). 

We can draw two conclusions 
here from Table~\ref{generalization}:
each attack model performs best within its distribution, and each attack model's performance generalizes well on unseen datasets (distributions). 
For example, the results of BDD100K and Mapillary 
show that both models trained on these two datasets are able
to reach comparable performance when generalizing to others.
We conjecture that a larger dataset may contain
more useful information during training, thus resulting in
better generalization ability.

\begin{table*}[!]
\centering
\begin{tabular}{lcccc}
 \toprule
 Model type &  Cityscapes & BDD100K &  Mapillary\\  
\midrule
 Normal    &  74.71\% &  66.56\%  & 68.39\%\\
 Regularizer-removed &  62.87\% &  54.88\%  & 59.98\%\\
 \bottomrule
\end{tabular}
 \caption{The success rate of our model (as shown in Figure~\ref{fig:structure}) with/without the proposed regularizer across different datasets.} \label{tab:regularizer} 
\end{table*}

\begin{table*}[!]
\centering
\begin{tabular}{lcccccc}
 \toprule
 attack model size & target model size & ratio & Cityscapes & BDD100K &  Mapillary\\  
\midrule
 Model\#1 (531M)  & 269M \iffalse134M\fi& 1.974 & 96.34\% &  97.86\%  & 96.09\% \\
 Model\#2 \iffalse(3.02M)\fi(10.69M)  & 269M \iffalse134M\fi& 0.040 & 74.71\% &   66.56\%  & 68.39\% \\
 Model\#3 \iffalse(1.52M)\fi(2.69M)  & 269M\iffalse134M\fi& 0.010 & 64.47\% &  53.63\%  & 60.39\% \\
 Model\#4 \iffalse(770K)\fi(692K)  & 269M\iffalse134M\fi&0.003 & 60.63\% &  50.99\%  & 45.53\%\\
 \bottomrule
\end{tabular}
 \caption{The parameter-wise efficiency of our model on type \#1 attack. Model \#1 adopt the generator in~\cite{gap}. Model \#2 uses the generator in Figure~\ref{fig:structure}. Model \#3 is the same as Model \#2 only the number of feature maps in each layer is cut in half. Similarly, Model \#4 cuts half of its parameters from Model \#3.} 
 \label{tab:param_efficiency}
\end{table*}

\subsection{Generalization across models} \label{en_across_models}
Generalization across models is also an important aspect of an adversarial model. We select two target models based on DeepLabV3Plus~\cite{deep_lab_v3_plus}. One has a backbone of MobileNets~\cite{mobile_nets}, while the other one has a backbone of ResNet. Since the target model is very deep, we replace the generator with a deeper version, a ResNet-based FCN. As Table~\ref{gen_across_models} shows, for all three attack types, the success rates are quite close when the two target models are switched.

\subsection{Ablation study}\label{ablation}
The regularizer in our model plays a vital role in attacking
segmentation models dynamically. The potential assumption
is that if a generator can create perturbations for dynamic
target labels, then the embedded features that come from the
generator should also have a good knowledge of the spatial 
structure within each image. As a result, we can use these 
embedded features to generate labels that come from the target model.
To explore the role of the regularizer, we compare the 
performance (manipulated rate) between our models and the 
corresponding regularizer-free ones on three datasets. 
As Table~\ref{tab:regularizer} shows, 
the regularizer can improve the manipulated rate by 
around 12\% in Cityscapes and Mapillary, and 8\% in Mapillary.

\subsection{Efficiency of attack models}~\label{para_efficiency}
The efficiency of attack models has been largely
ignored by previous works. However, as attack/target
models become larger and more complex, the efficiency issue
can no longer be ignored.  For example, the success rate is higher when using a ResNet-based generator to attack a MobileNets-based target model compared to vice versa.

Here we propose a metric that is defined as the ratio of attack models' sizes and the target models' sizes. For the same manipulated rate, smaller attack models are more efficient.  We list four different models and show the total number of parameters of them
in Table~\ref{tab:param_efficiency} in descending order. 

We can see that larger attack models can achieve a higher success rate. For example, when the model is twice as large as the target model, model \#1 can achieve an almost perfect manipulated rate (more than 96\%). With only 4\% parameters, our model \#2 can achieve around 70\% manipulated rate. It is interesting that with less than 0.3\% parameters (model \#4), our model can still achieve relatively good performance.

\section{Conclusion}
We propose a framework for creating semantically stealthy adversarial attacks against segmentation models. In particular, we design an algorithm that can manipulate targeted pixels' labels as designed while keeping  other pixels' labels untouched. To achieve this goal, we introduce prior knowledge (each perturbation is conditionally dependent on the corresponding input image) as well as a special regularizer into our attack model. We evaluate our model's performance by success rates on three types of stealthy attacks. The experiment shows that our framework has a relatively high success rate across datasets/models. Finally, we propose the concept of attack efficiency which may help estimate attack models' parameter-wise efficiency. 

%-------------------------------------------------------------------------

{\small
\bibliographystyle{plainnat}
\bibliography{egbib}
}
\begin{figure*}[h]
\begin{center}
    \includegraphics[width=0.20\linewidth]{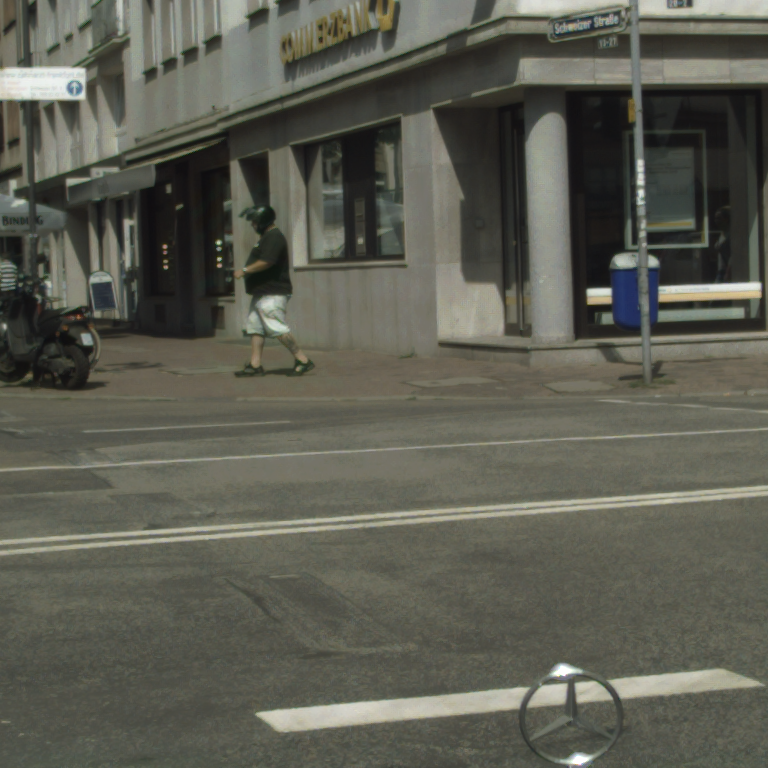}
    \includegraphics[width=0.20\linewidth]{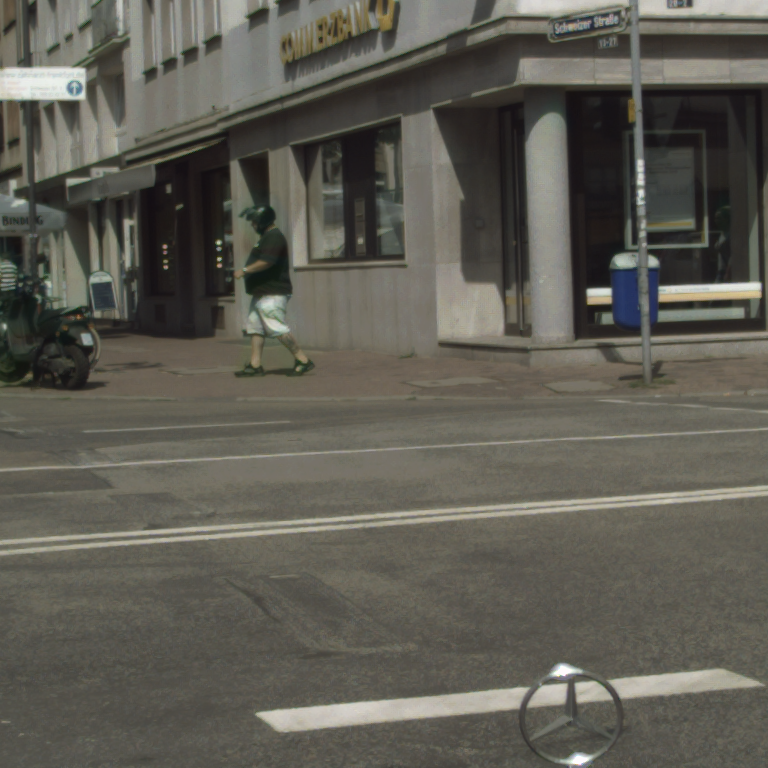}
    \includegraphics[width=0.20\linewidth]{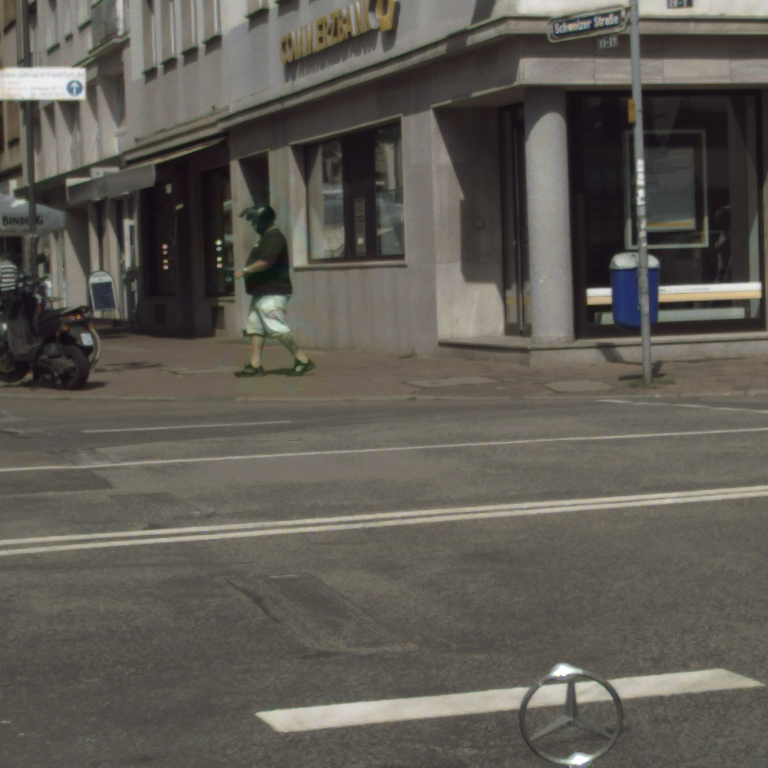}
    \includegraphics[width=0.20\linewidth]{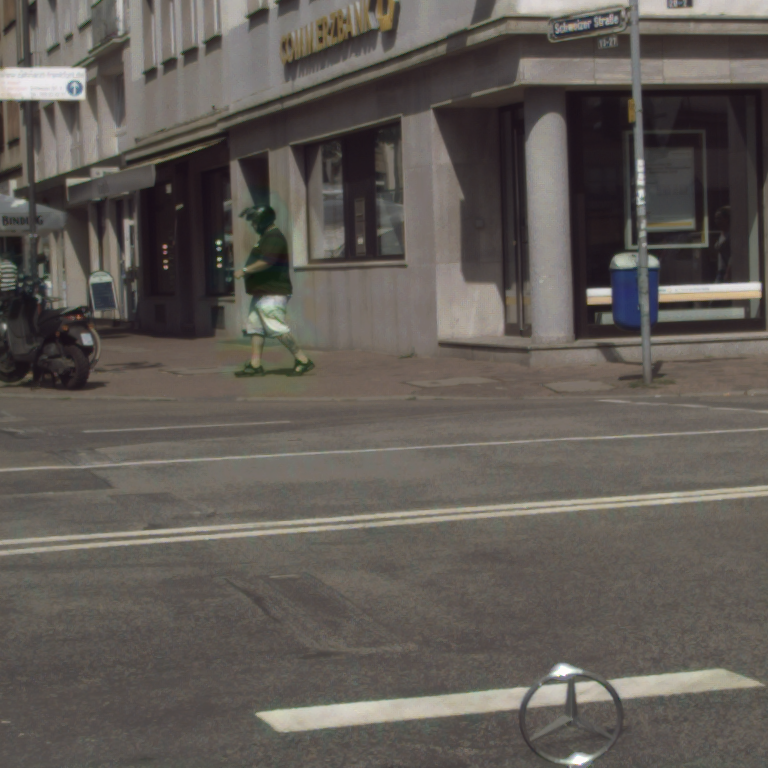}\\
    \includegraphics[width=0.20\linewidth]{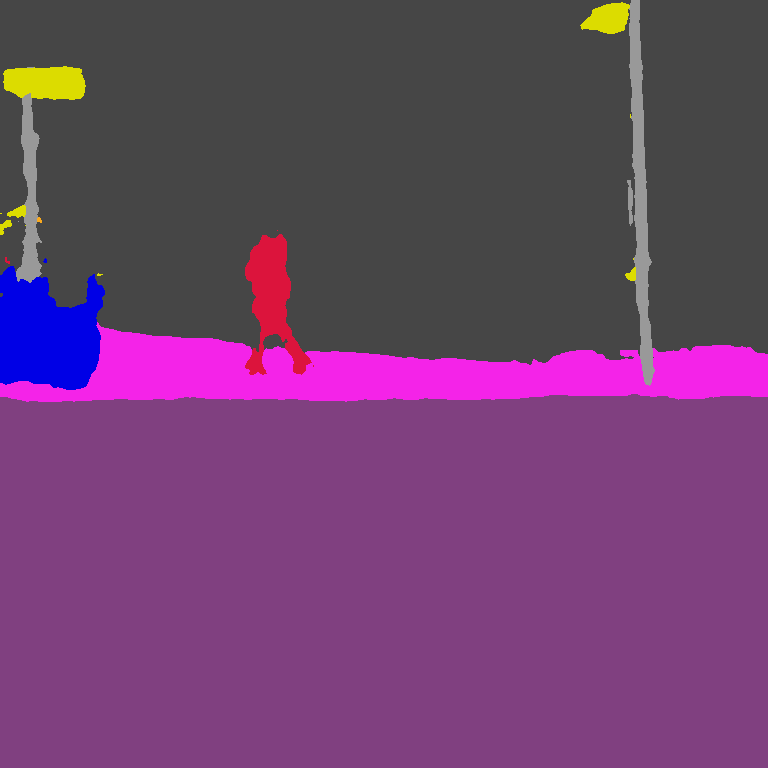}
    \includegraphics[width=0.20\linewidth]{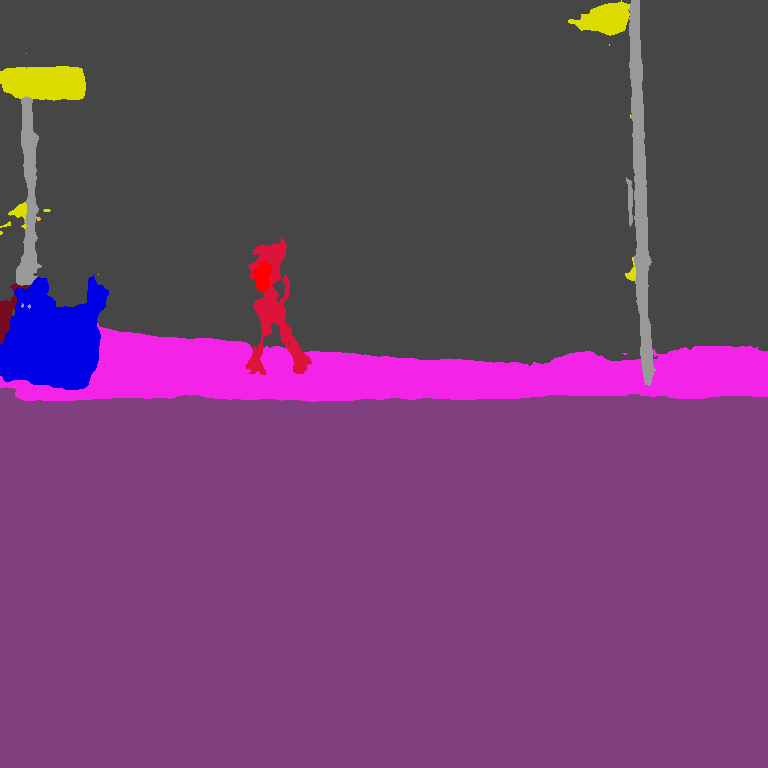}
    \includegraphics[width=0.20\linewidth]{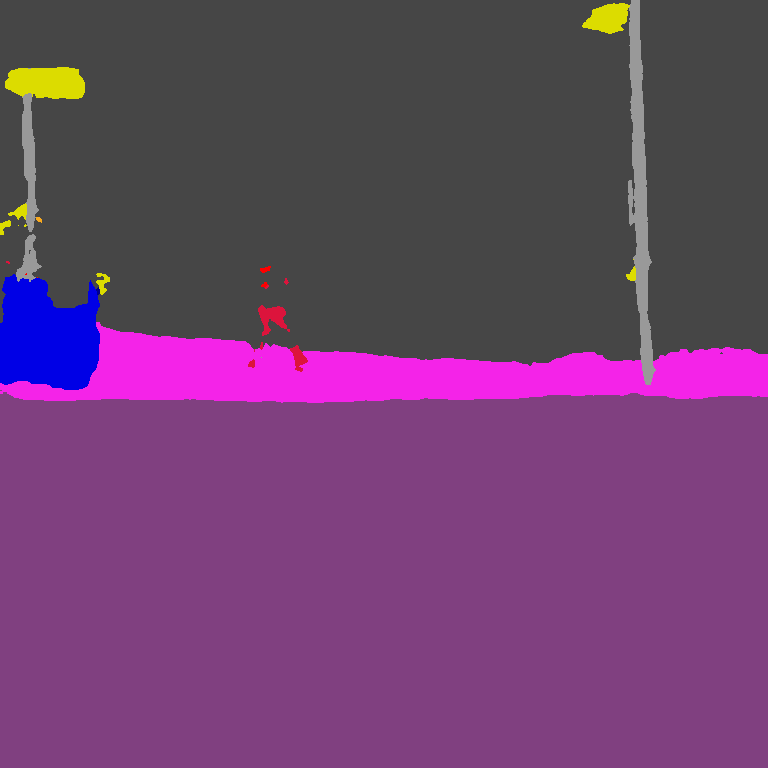}
    \includegraphics[width=0.20\linewidth]{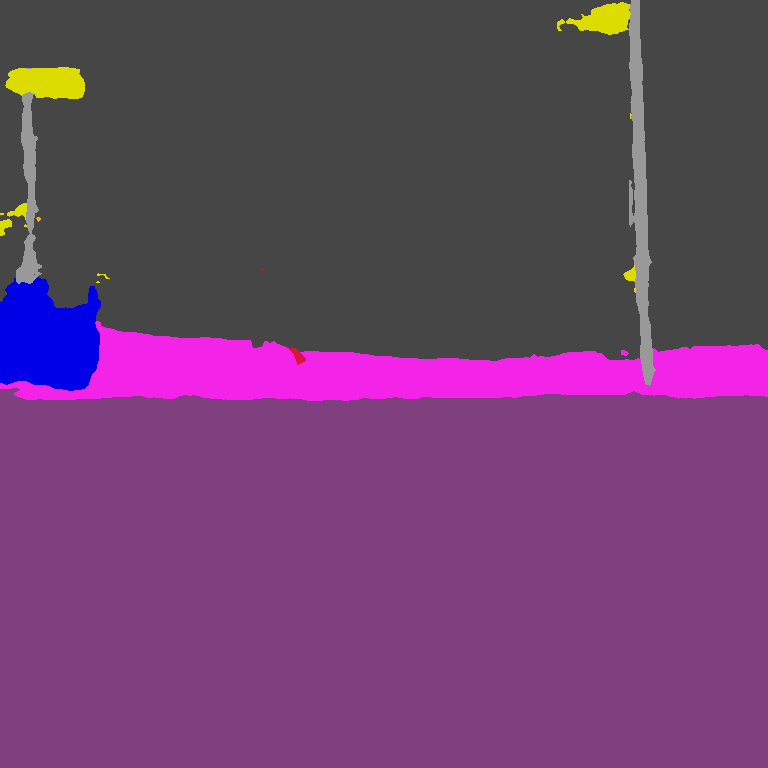}
    
\end{center}
  \caption{The relation between infinity norm and the corresponding type\#1 performance. From left to right, the infinity norm is 4, 6, 8, 10 separately.}
\label{fig:inf_norm}
\end{figure*}

\begin{figure}[t]
\begin{center}
    \includegraphics[width=0.48\linewidth]{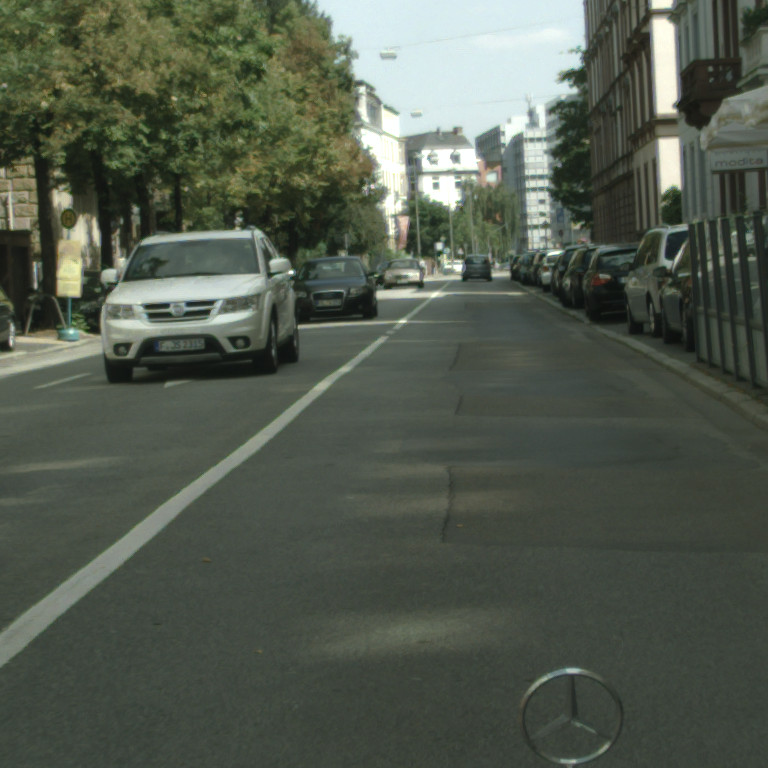}
    \includegraphics[width=0.48\linewidth]{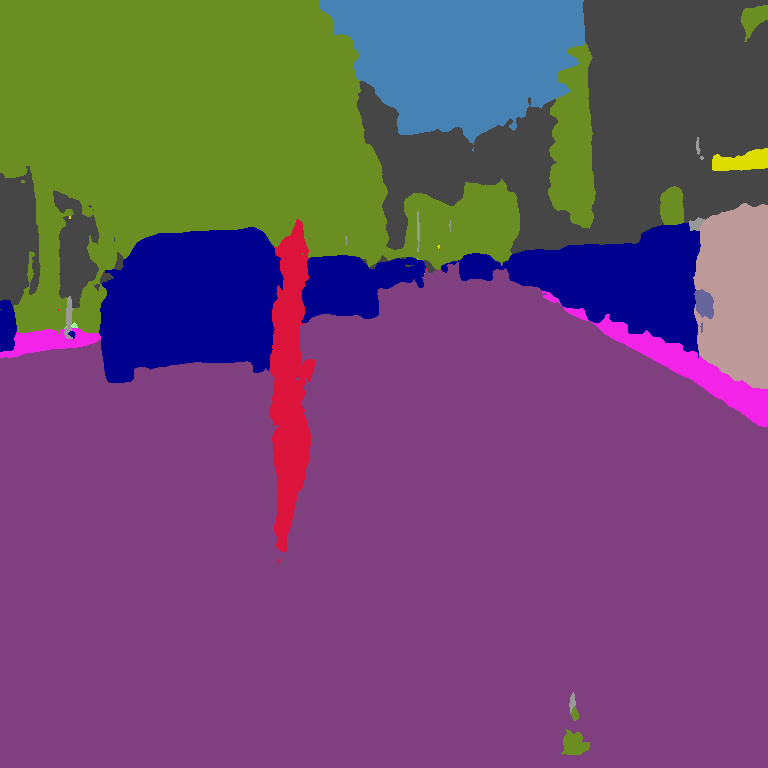}\\
    \includegraphics[width=0.48\linewidth]{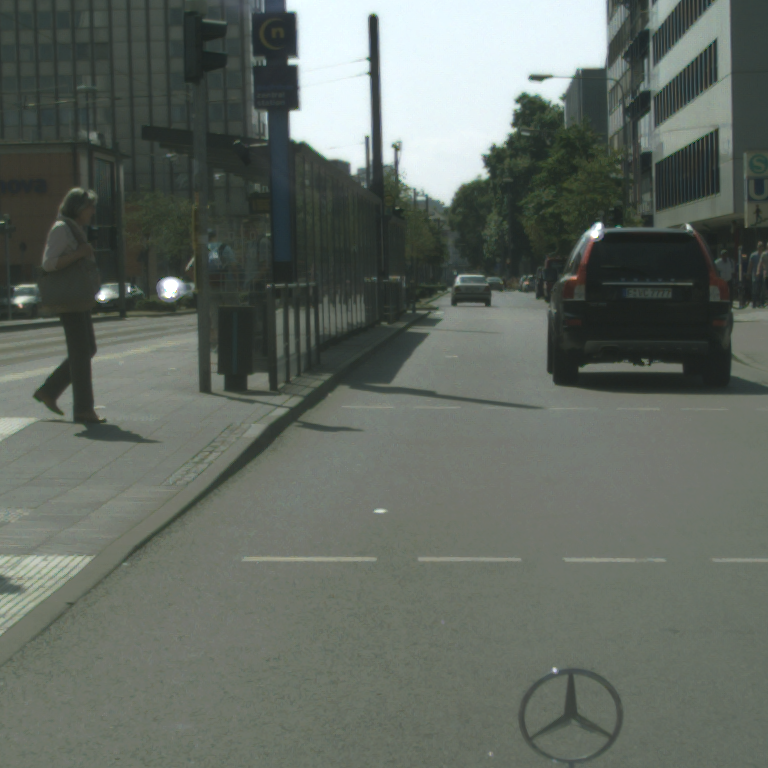}
    \includegraphics[width=0.48\linewidth]{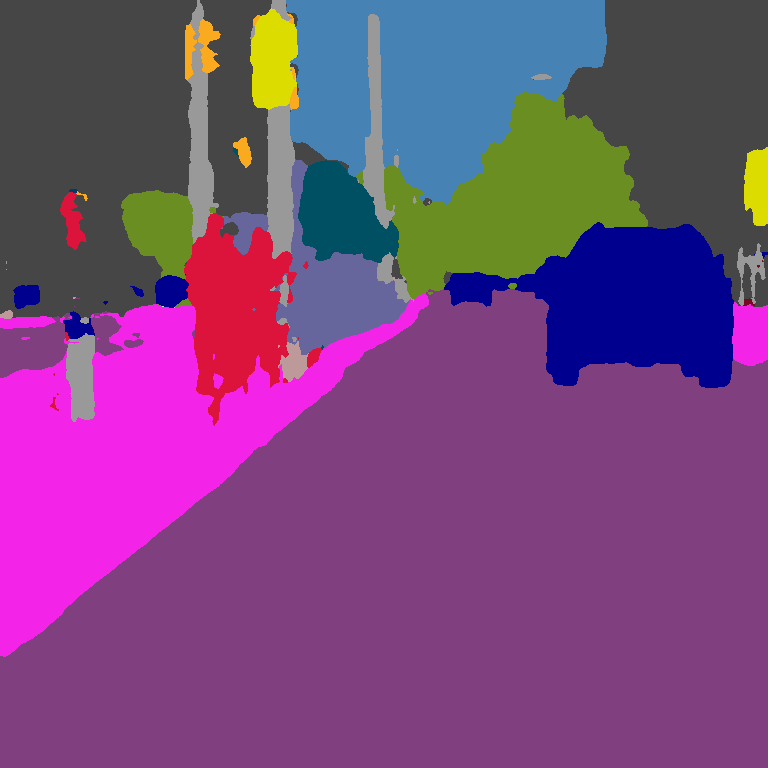}
\end{center}
  \caption{CityScapes Type\#2 and Type\#3 attack on DeepLabV3Plus model with MobileNet backbone. For Type\#2 attack, the model learns to add a new set of 'person' labels in the prediction. Type\#3 attack combines both Type\#1 and Type\#2 by vanishing `person' and `rider' labels into the background, and adding a new set of 'person' labels in the prediction. Top left: The input image. Top right: The input image + perturbations. Lower left: Predictions before attacks. Lower middle: Manipulated label mask. Lower right: Predictions after attacks.}
\label{fig:mobile_type2_type3}
\end{figure}

\pagebreak

\input{appendix}

\end{document}

%% file: math_commands.tex
\usepackage{amsmath,amsfonts,bm}

\def\eqref#1{equation~\ref{#1}}
\def\1{\bm{1}}

\DeclareMathAlphabet{\mathsfit}{\encodingdefault}{\sfdefault}{m}{sl}
\SetMathAlphabet{\mathsfit}{bold}{\encodingdefault}{\sfdefault}{bx}{n}

\def\gD{{\mathcal{D}}}

\def\gL{{\mathcal{L}}}
\def\gM{{\mathcal{M}}}

\newcommand{\R}{\mathbb{R}}

%% file: appendix.tex
\section{Supplementary Material}

\subsection{Infinity norm of the perturbations}
Although there have been lots of discussions on how the infinity norm affects the performance of adversarial attacks statistically, few of them are visualized. Here we select four infinity norm thresholds and visualize the corresponding performance on the same testing image. We choose to use DeepLabV3Plus-MobileNet as the target model. Figure~\ref{fig:inf_norm} shows how different infinity norms, 4, 6, 8, 10 affect the performance of `person' label vanishment. The results suggest that as the infinity norm increases, our model yields better performance against the target model.

\subsection{Results of DeepLabV3Plus on CityScapes}
\begin{figure*}[t]
\begin{center}
    \includegraphics[width=0.3\linewidth]{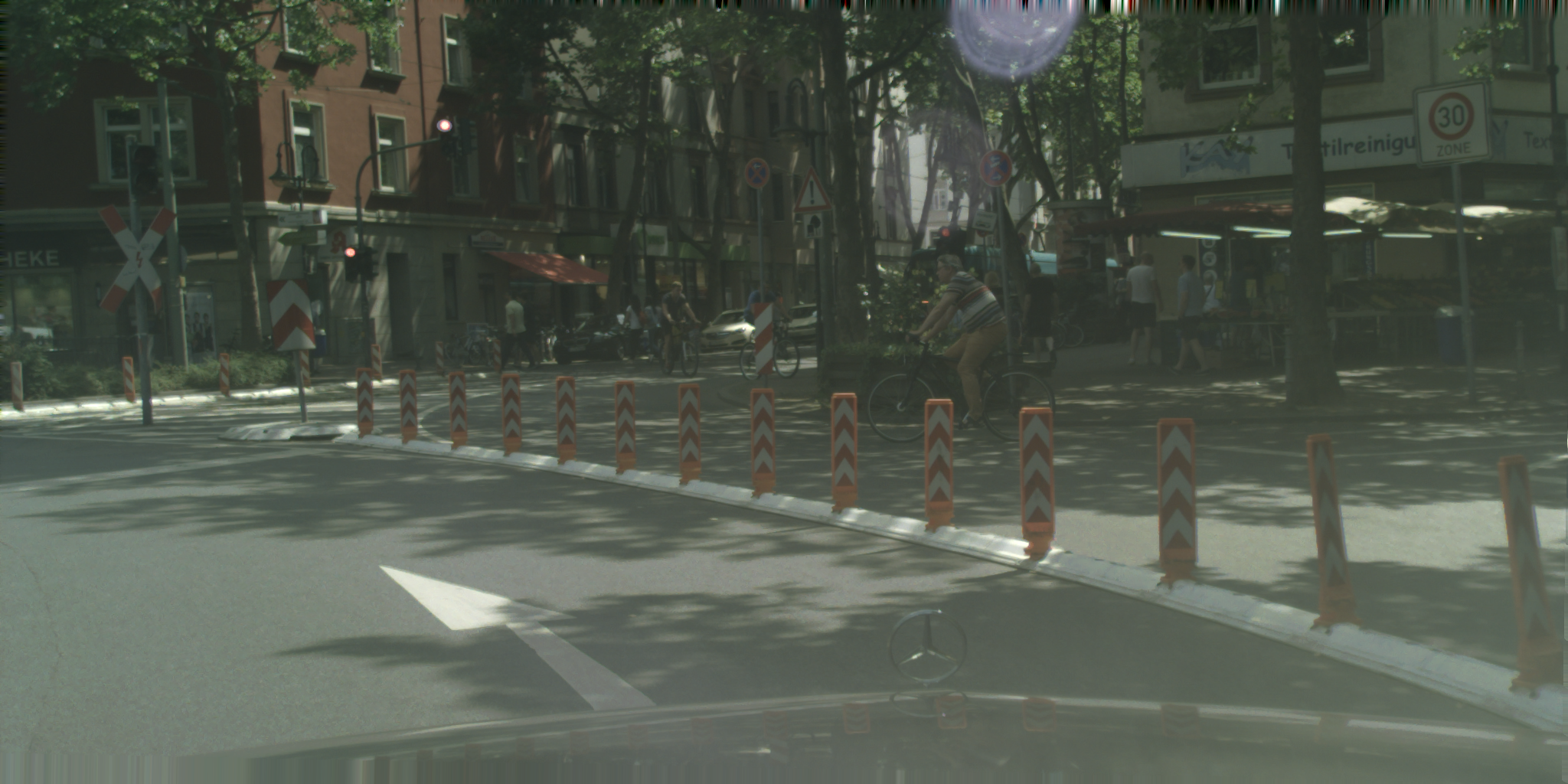}
    \includegraphics[width=0.3\linewidth]{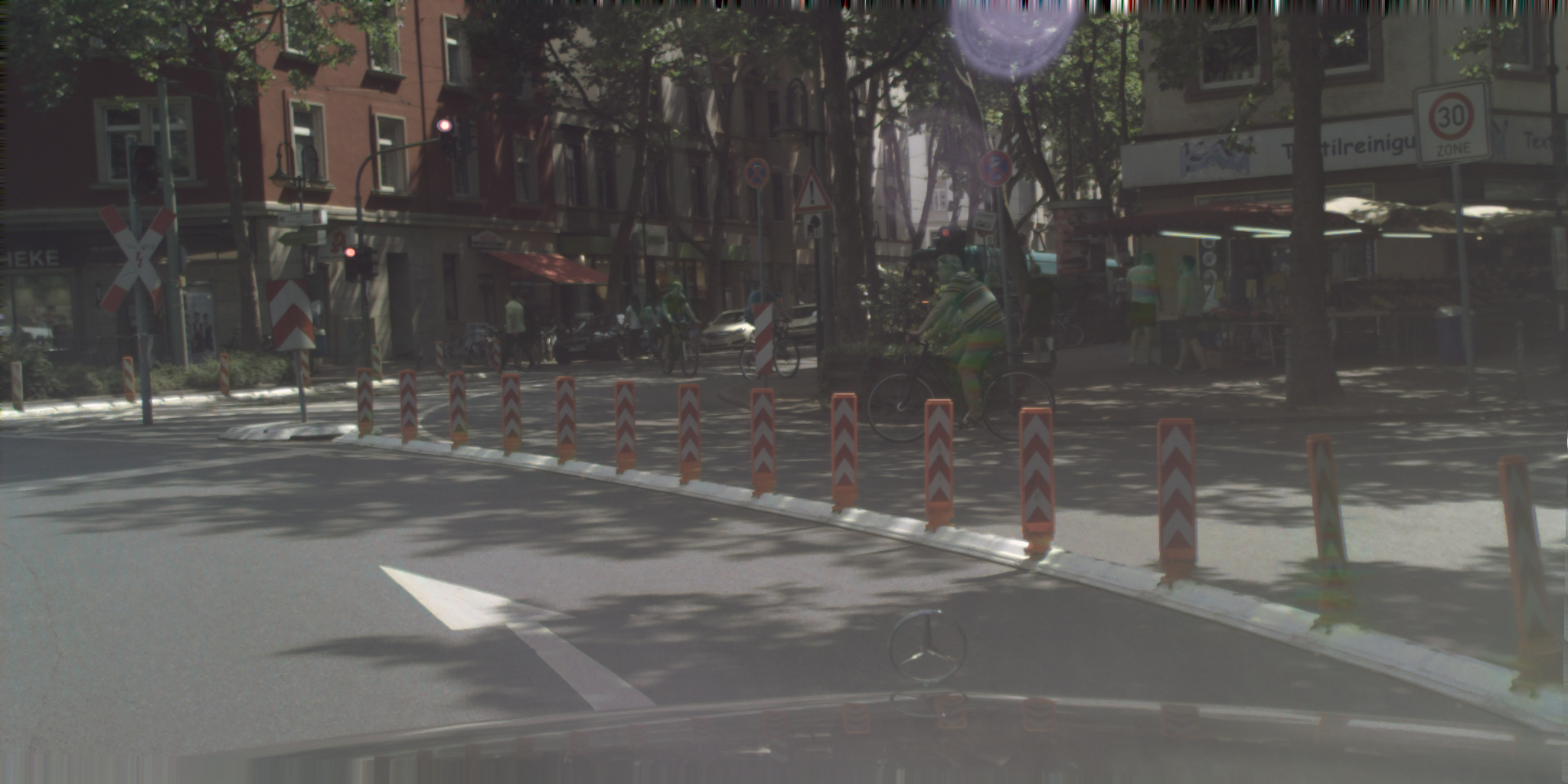}\\
    \includegraphics[width=0.3\linewidth]{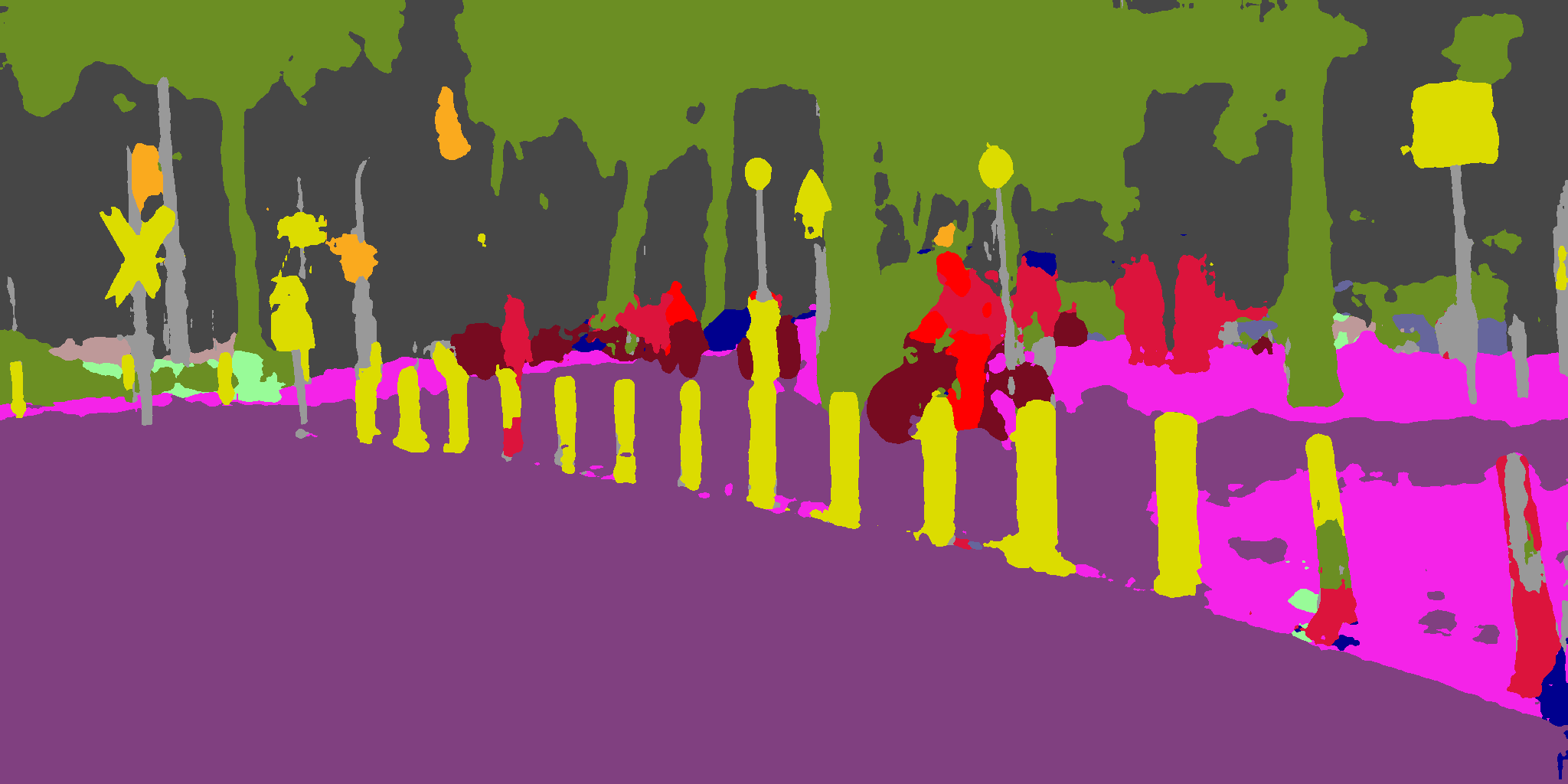}
    \includegraphics[width=0.3\linewidth]{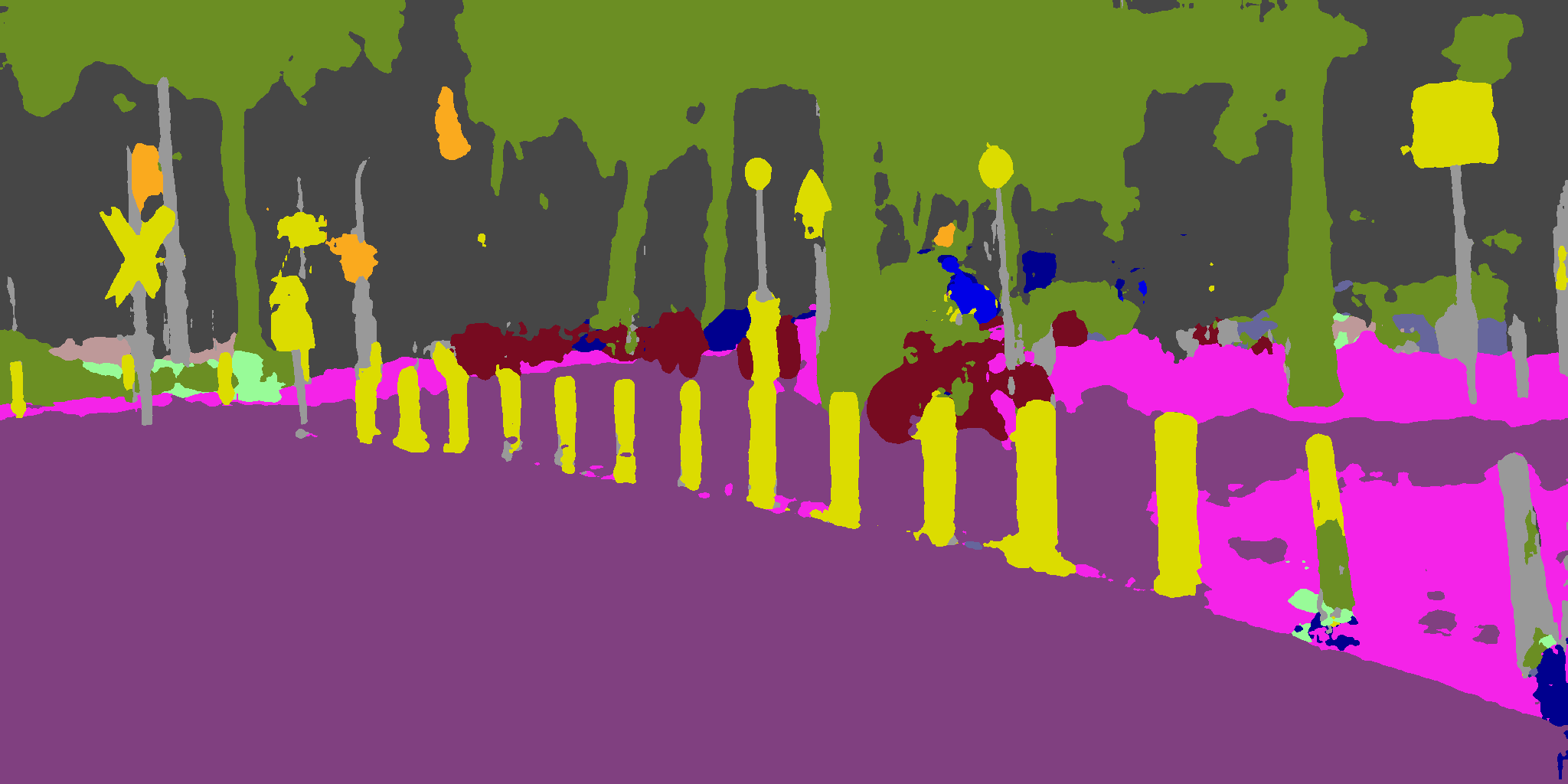}
    \includegraphics[width=0.3\linewidth]{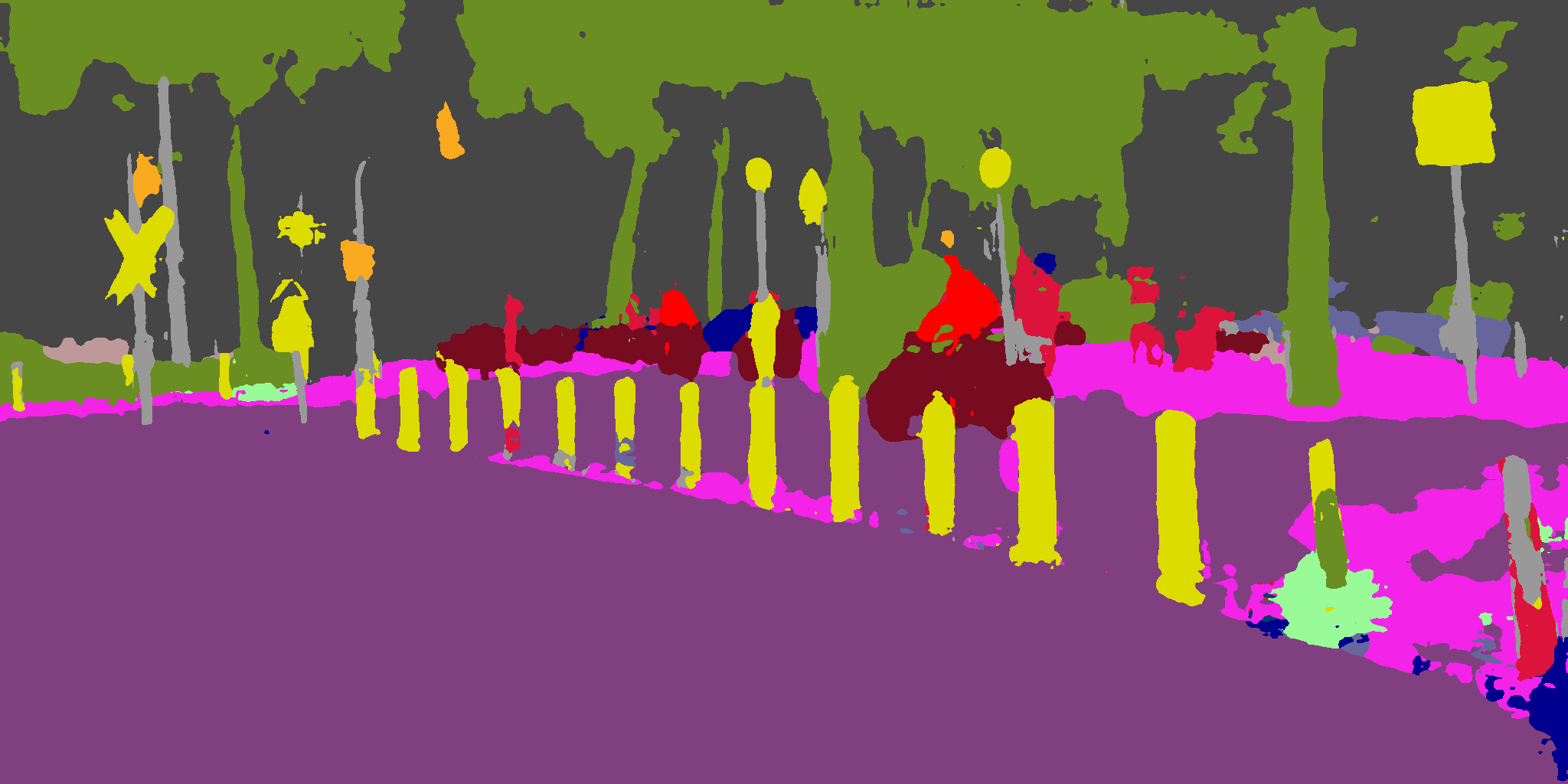}
\end{center}
   \caption{Type\#1 attack against DeepLabV3Plus-MobileNet model on Cityscapes. The `person' and 'rider' labels vanish into the background. Top left: The input image. Top right: The input image + perturbations. Lower left: Predictions before attacks. Lower middle: Manipulated label mask. Lower right: Predictions after attacks.}
\label{fig:trained_on_mobile_test_on_mobile}
\end{figure*}

\begin{figure*}[t]
\begin{center}
    \includegraphics[width=0.3\linewidth]{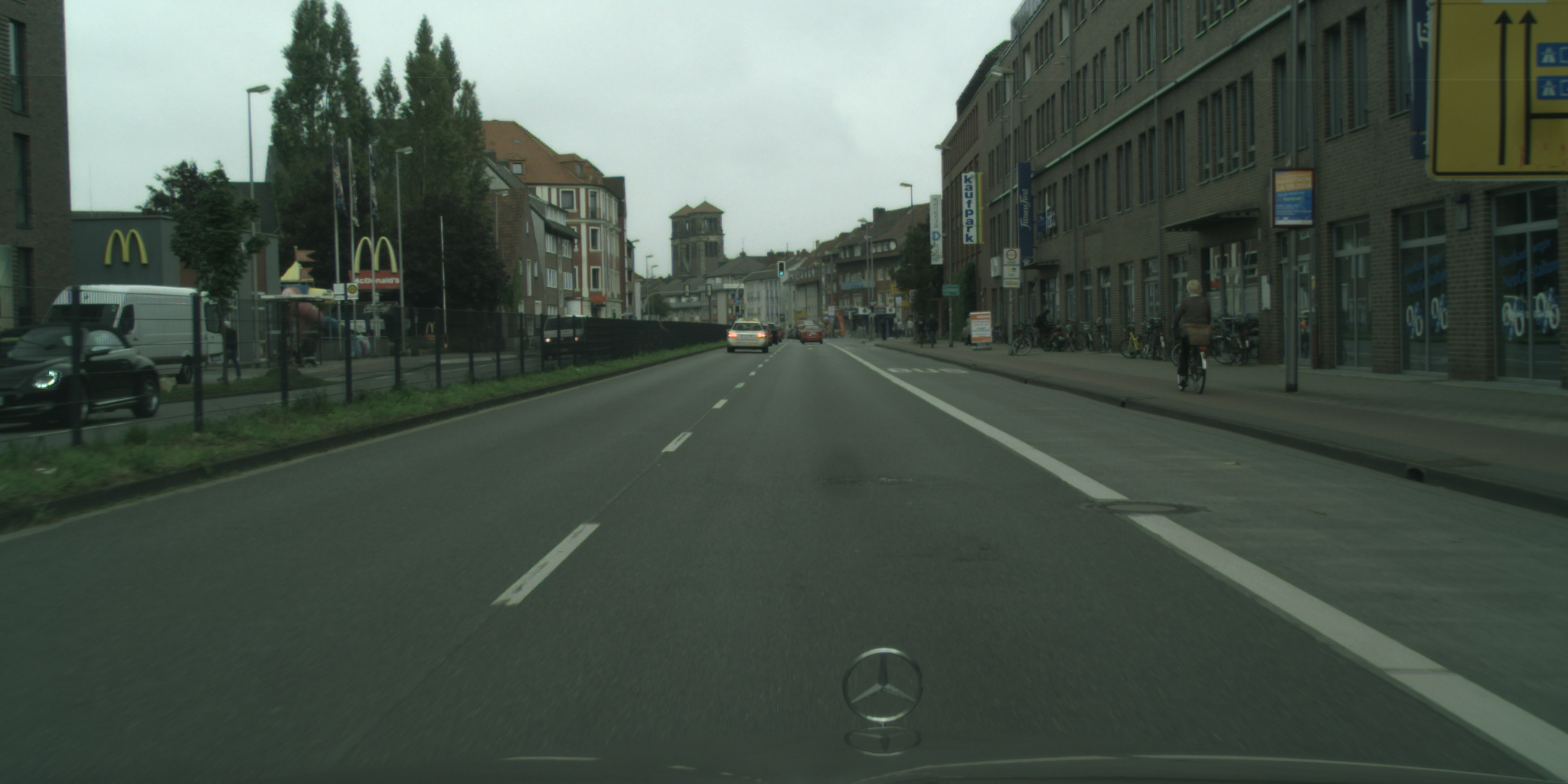}
    \includegraphics[width=0.3\linewidth]{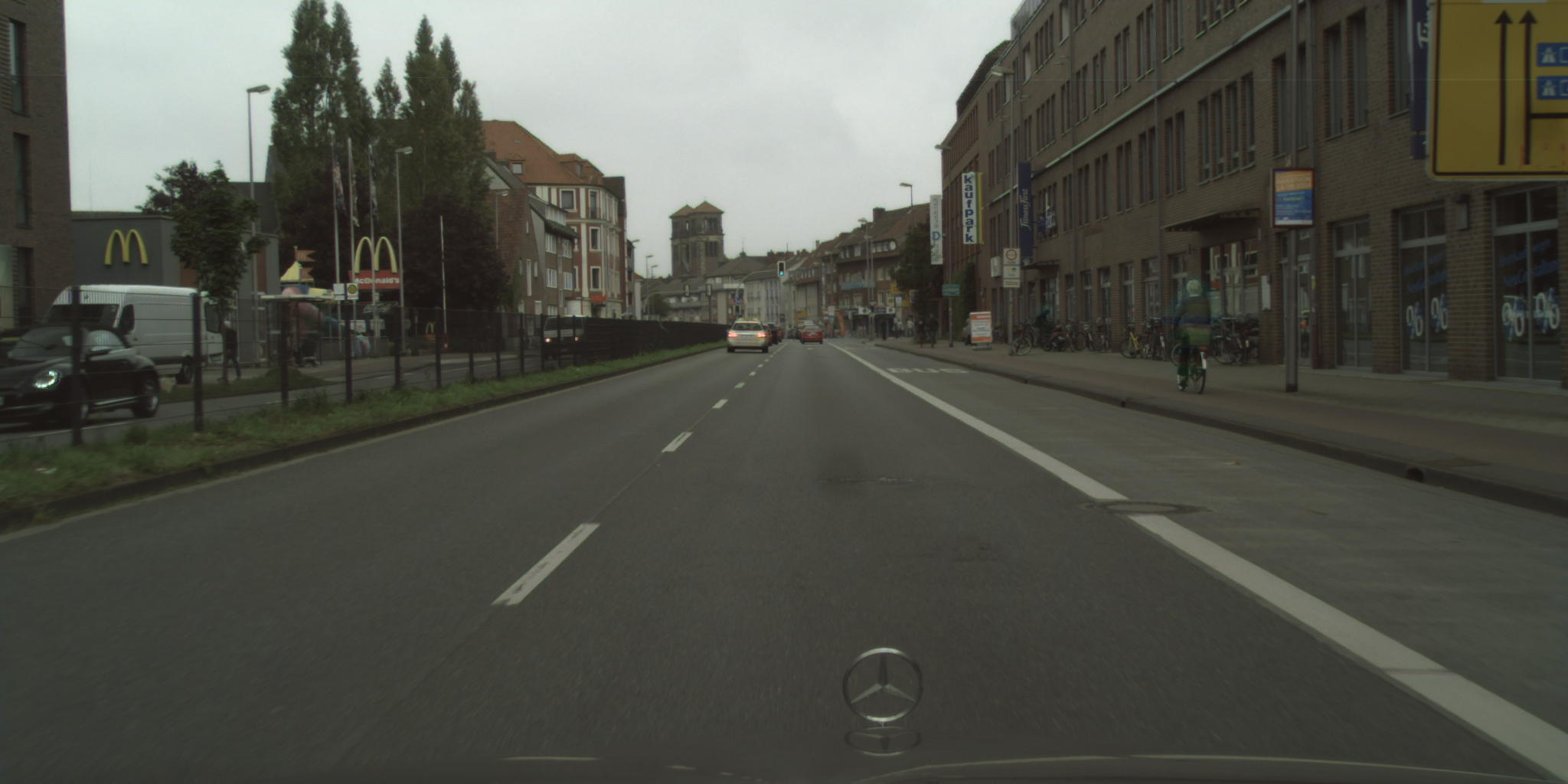}\\
    \includegraphics[width=0.3\linewidth]{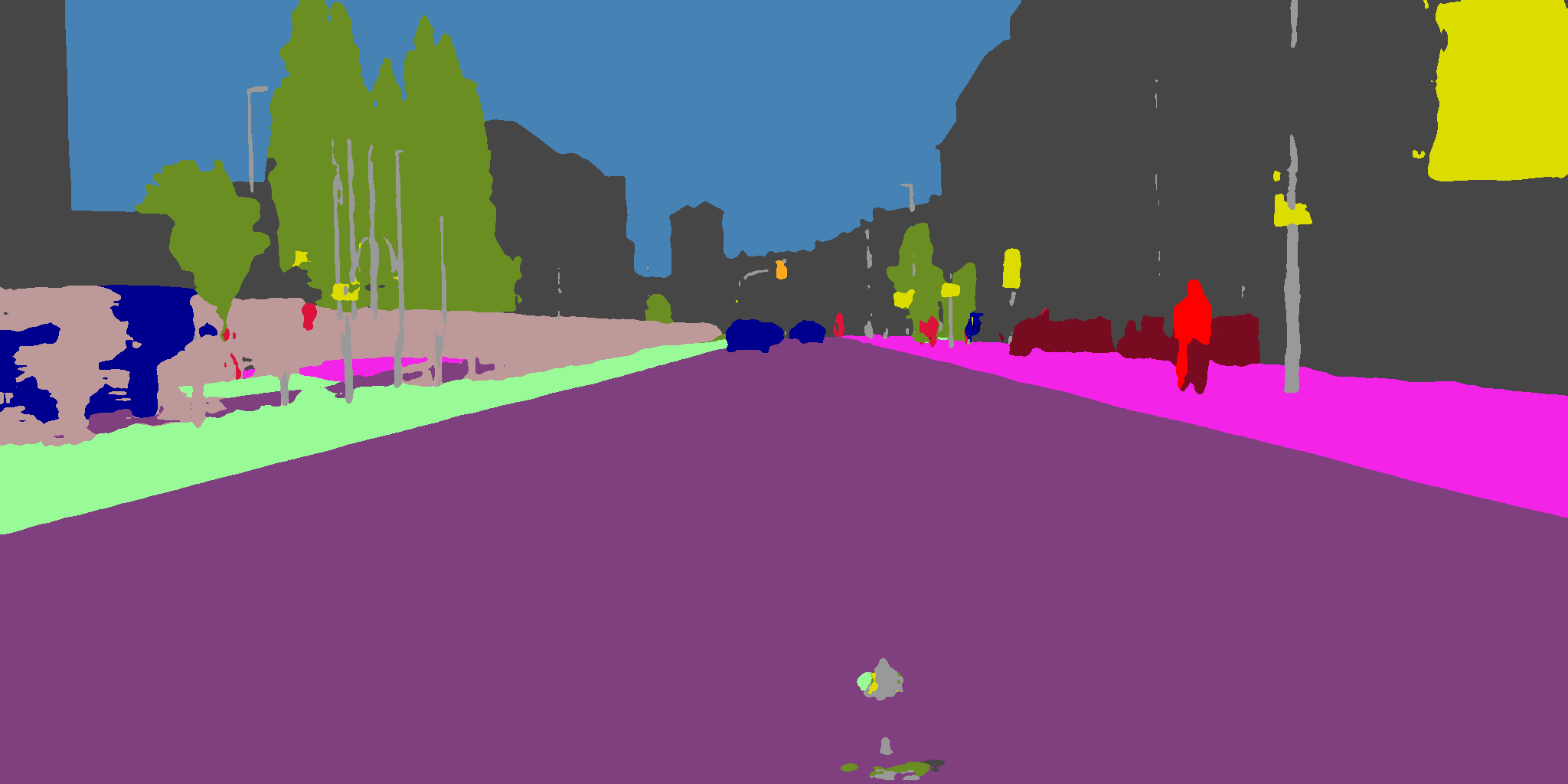}
    \includegraphics[width=0.3\linewidth]{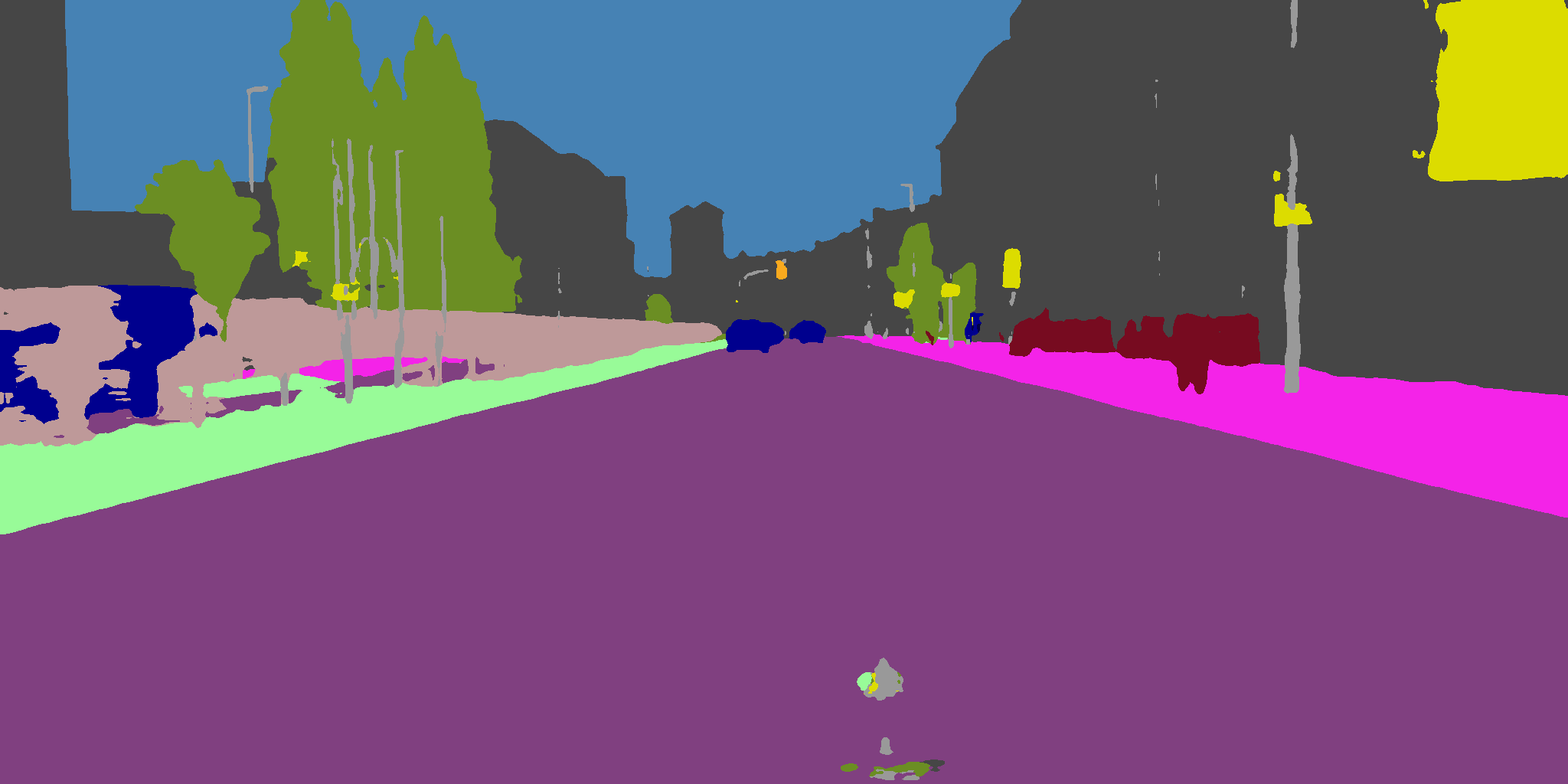}
    \includegraphics[width=0.3\linewidth]{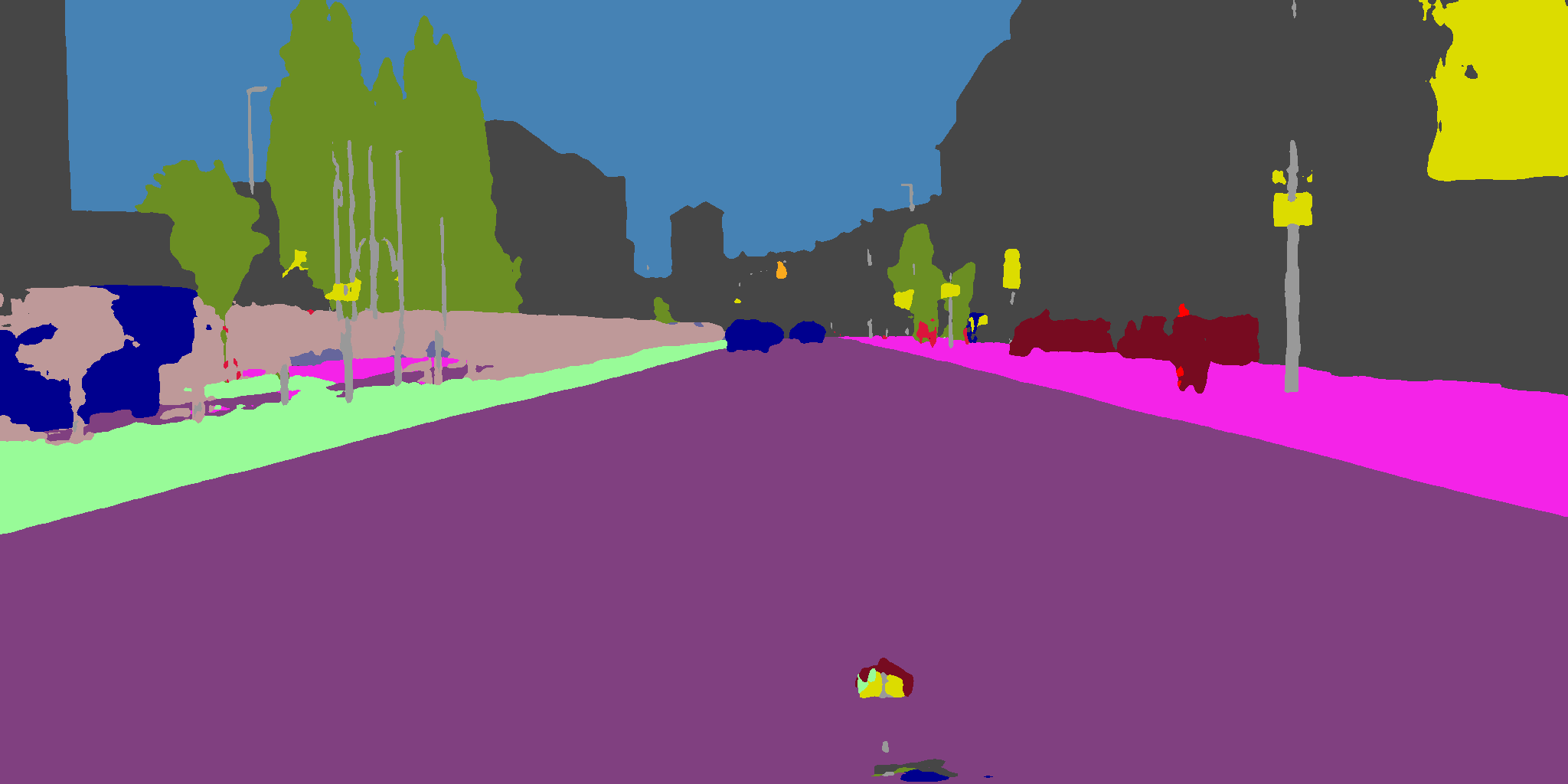}
\end{center}
   \caption{Type\#1 attack against DeepLabV3Plus-ResNet on Cityscapes. The `person' and `rider' labels vanish into the background. Top left: The input image. Top right: The input image + perturbations. Lower left: Predictions before attacks. Lower middle: Manipulated label mask. Lower right: Predictions after attacks.}
\label{fig:trained_on_resnet_test_on_resnet}
\end{figure*}

In this session, we visualize samples of the ResNet-based models attack against the DeepLabV3Plus-MobileNet-based target for different attack types. Figure~\ref{fig:trained_on_mobile_test_on_mobile}, Figure~\ref{fig:trained_on_resnet_test_on_resnet}, Figure~\ref{fig:mobile_type2_type3} show the result for type\#1, type\#2, type\#3, respectively. 

\begin{figure*}[t]
\begin{center}
    \includegraphics[width=0.3\linewidth]{images/resnet_mobilenet_10_self/0_image.png}
    \includegraphics[width=0.3\linewidth]{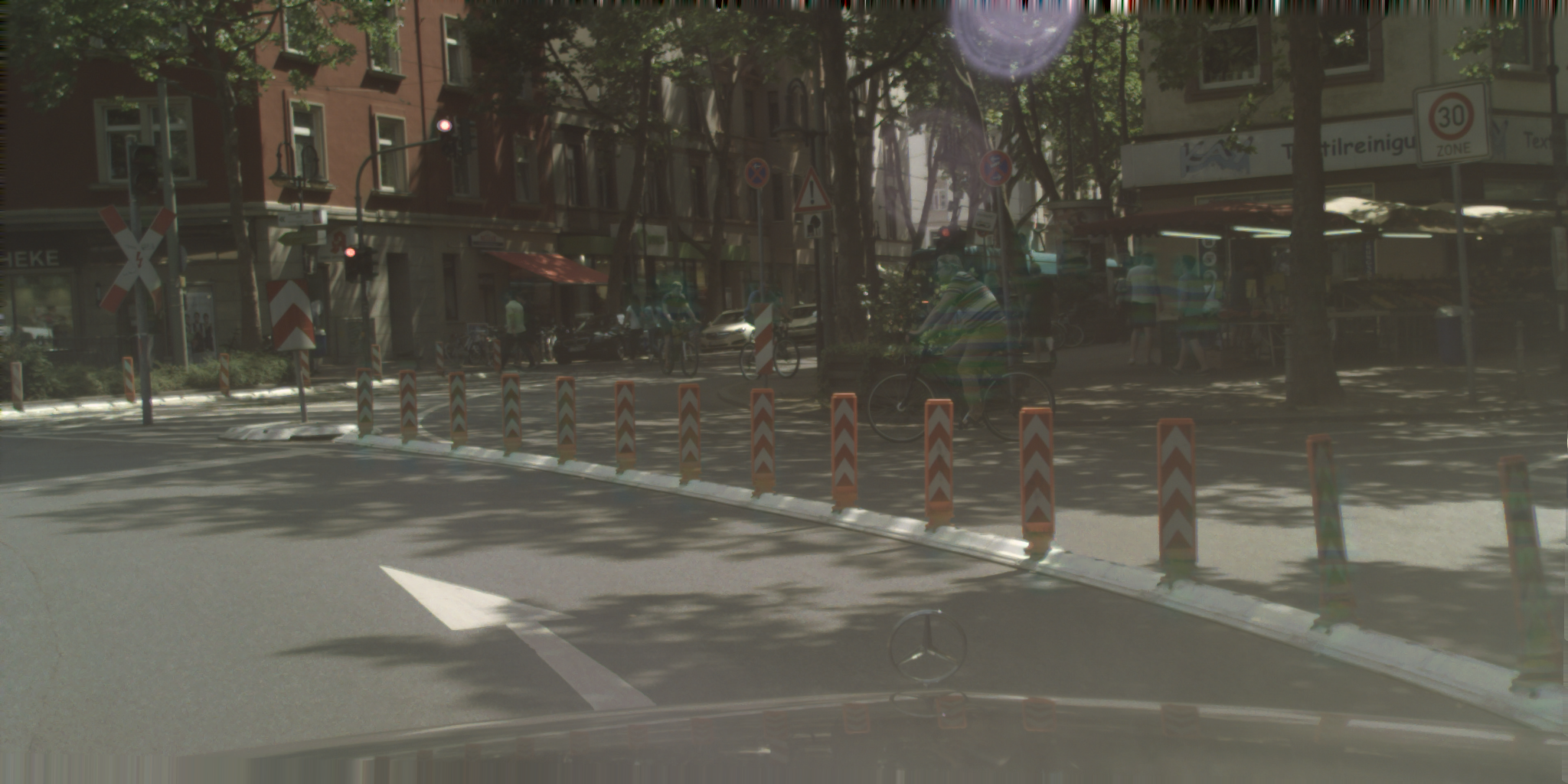}\\
    \includegraphics[width=0.3\linewidth]{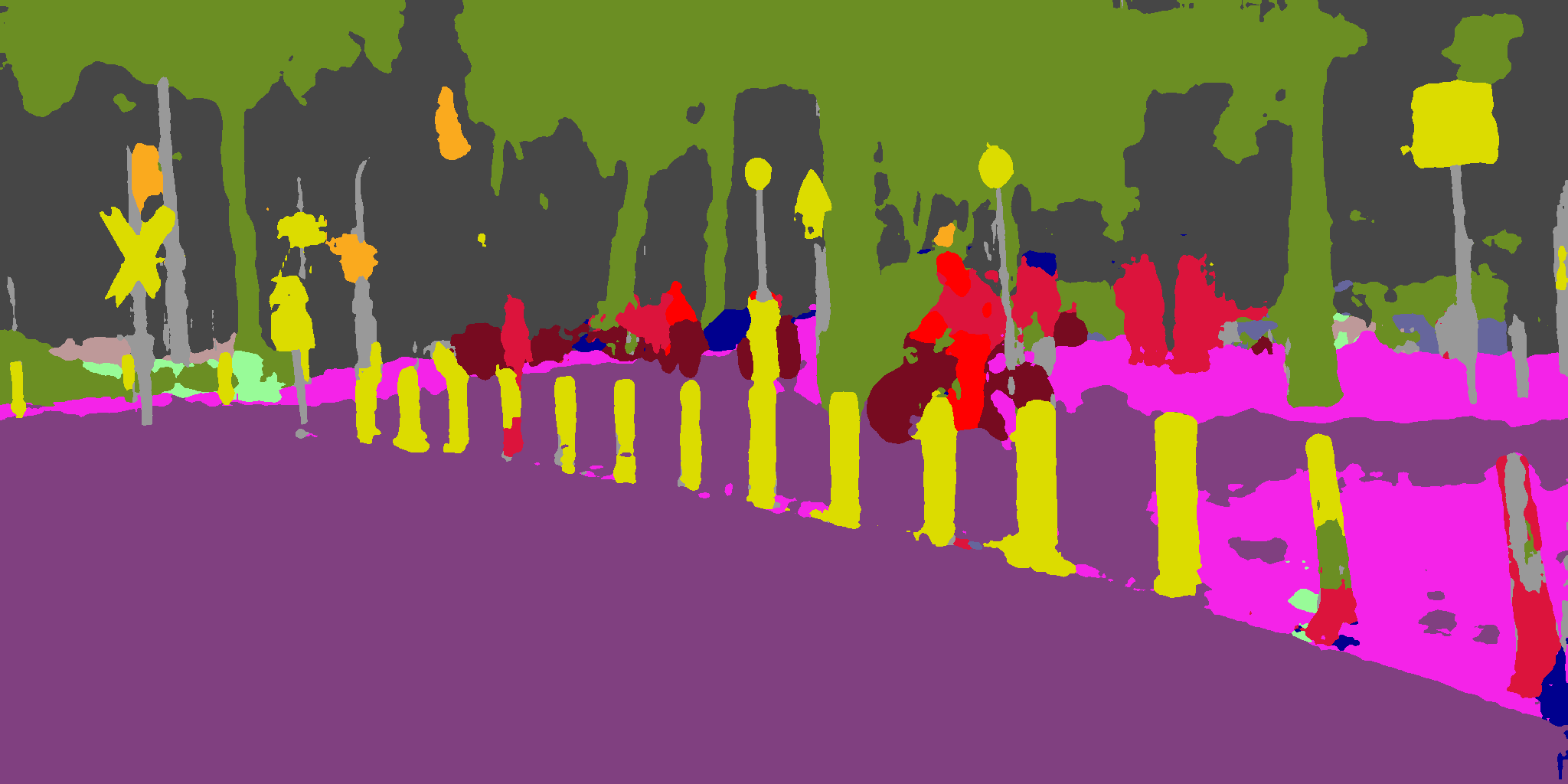}
    \includegraphics[width=0.3\linewidth]{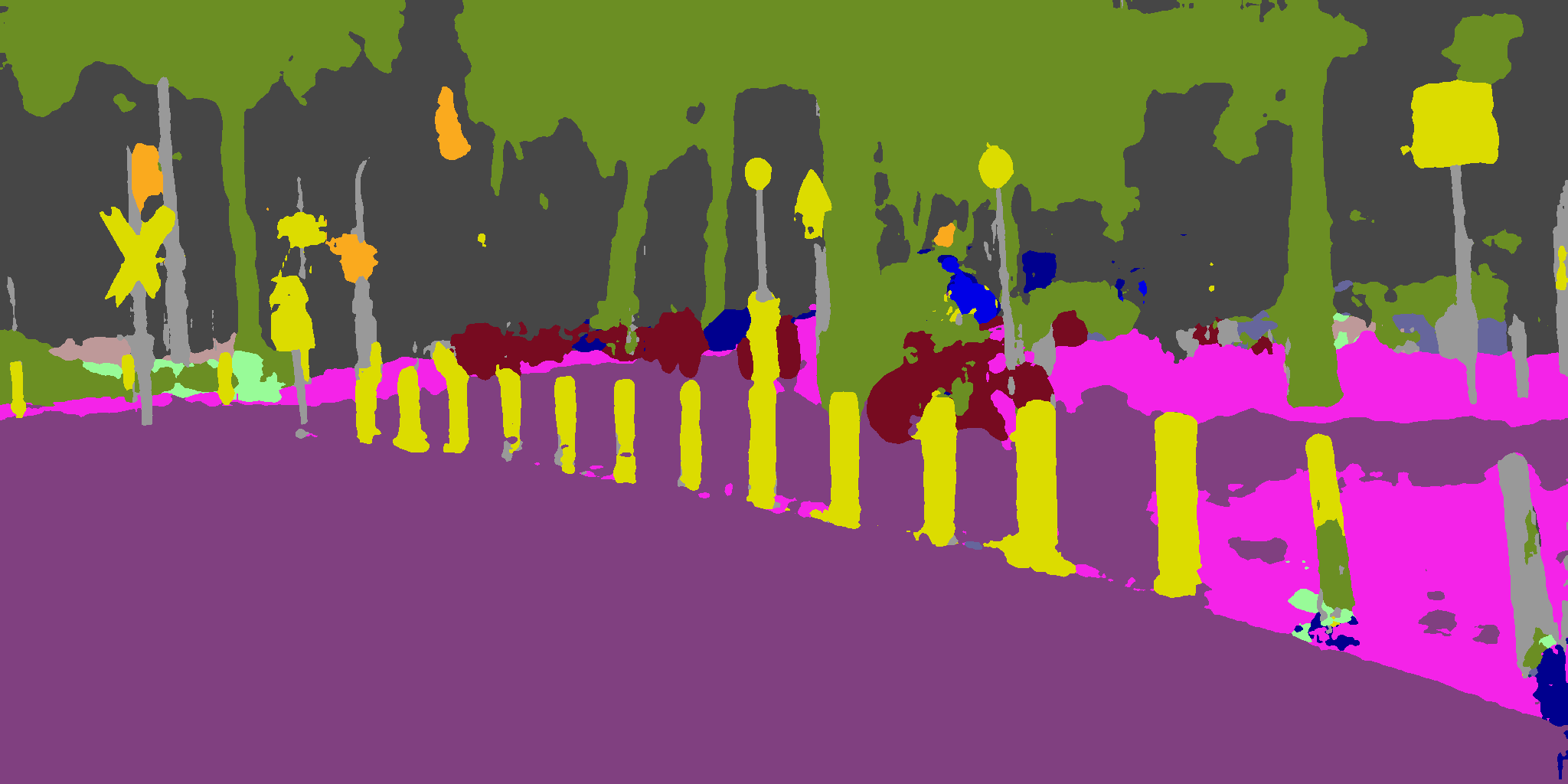}
    \includegraphics[width=0.3\linewidth]{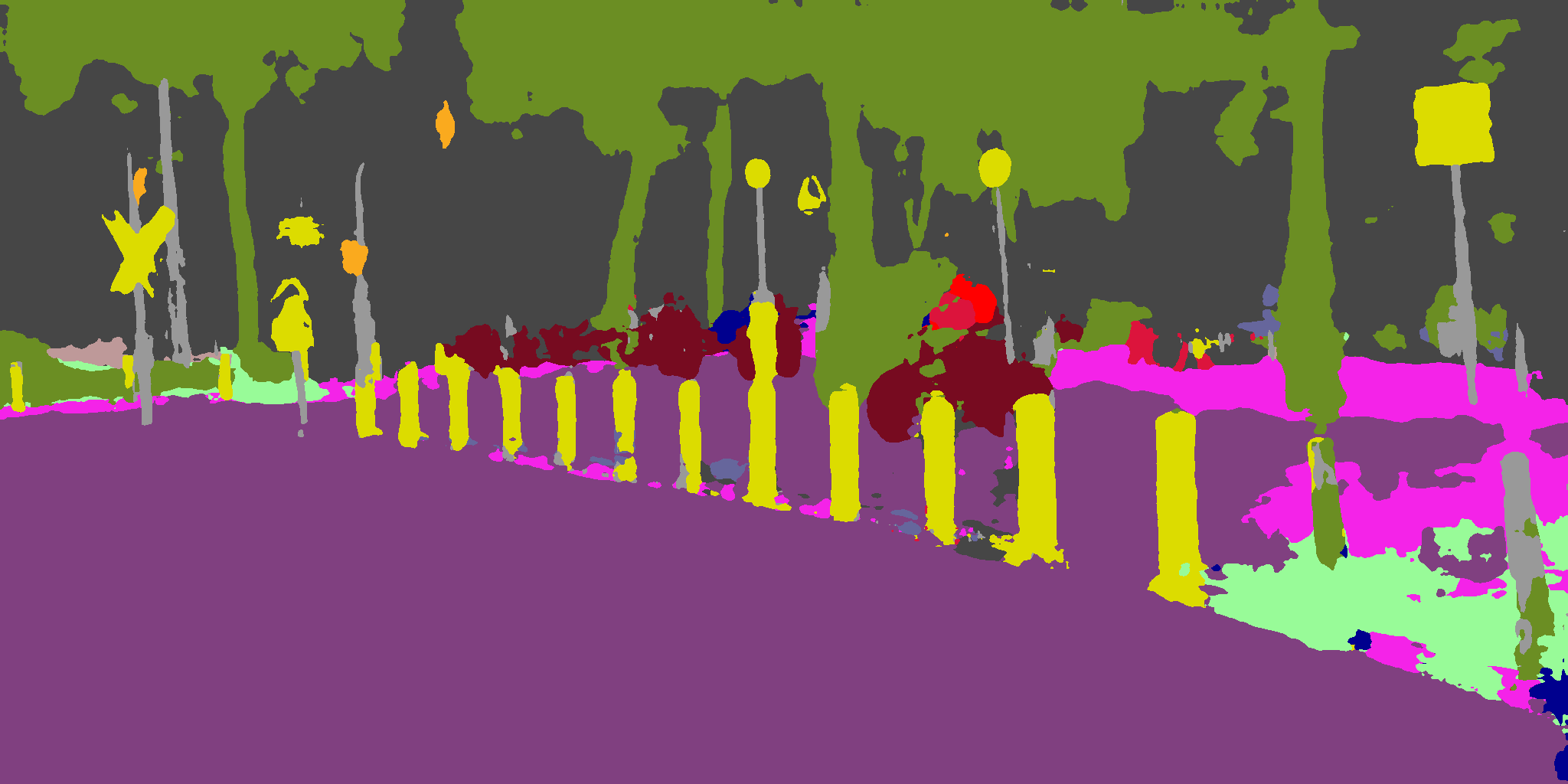}
\end{center}
   \caption{Type\#1 attack against DeepLabV3Plus-MobileNet on Cityscapes. The `person' and `rider' labels vanish into the background. Our model is trained to attack DeepLabV3Plus-Resnet and evaluate on DeepLabV3Plus-MobileNet. Top left: The input image. Top right: The input image + perturbations. Lower left: Predictions before attacks. Lower middle: Manipulated label mask. Lower right: Predictions after attacks.}
\label{fig:trained_on_resnet_test_on_mobile}
\end{figure*}

\begin{figure*}[t]
\begin{center}
    \includegraphics[width=0.3\linewidth]{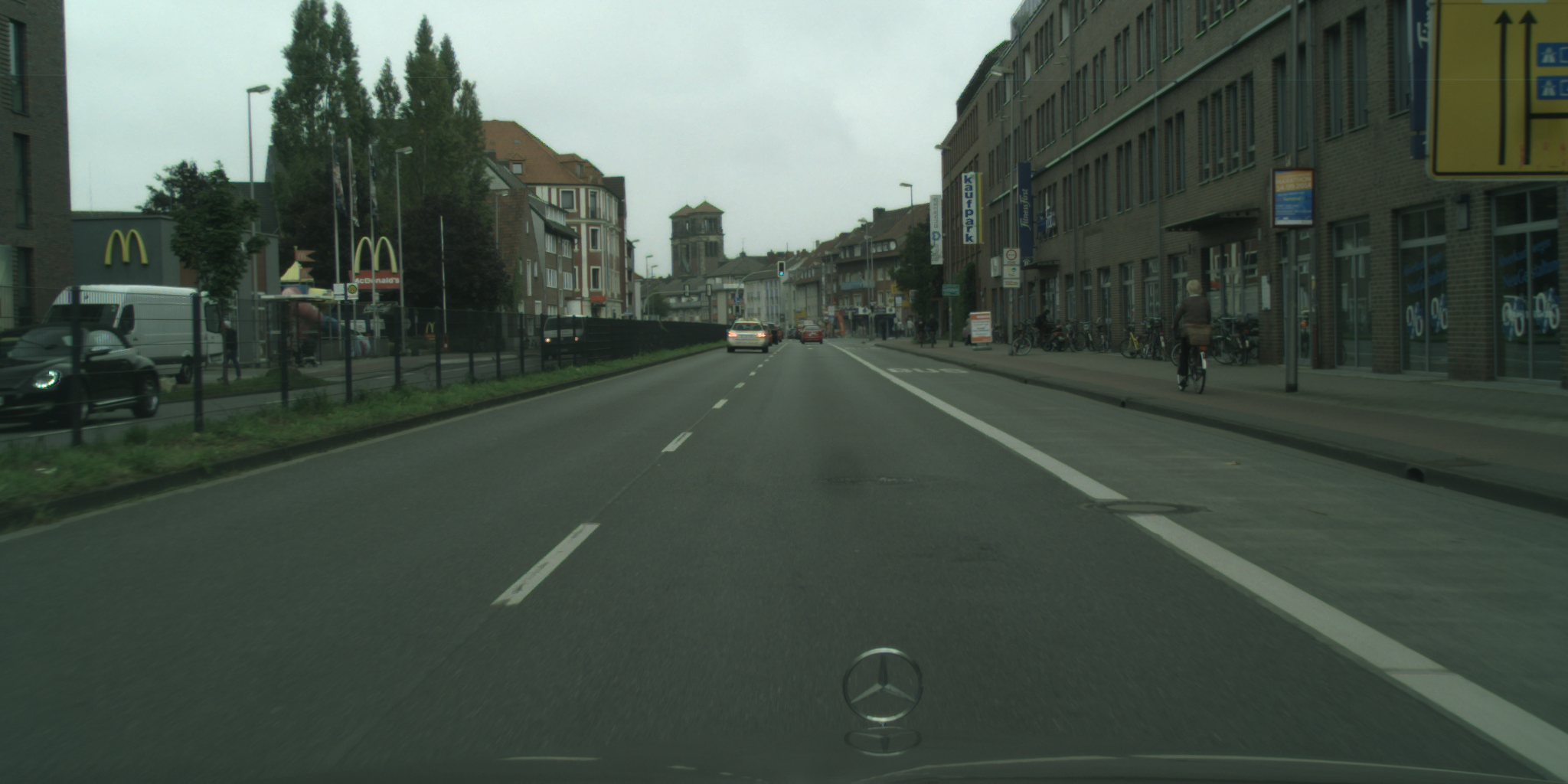}
    \includegraphics[width=0.3\linewidth]{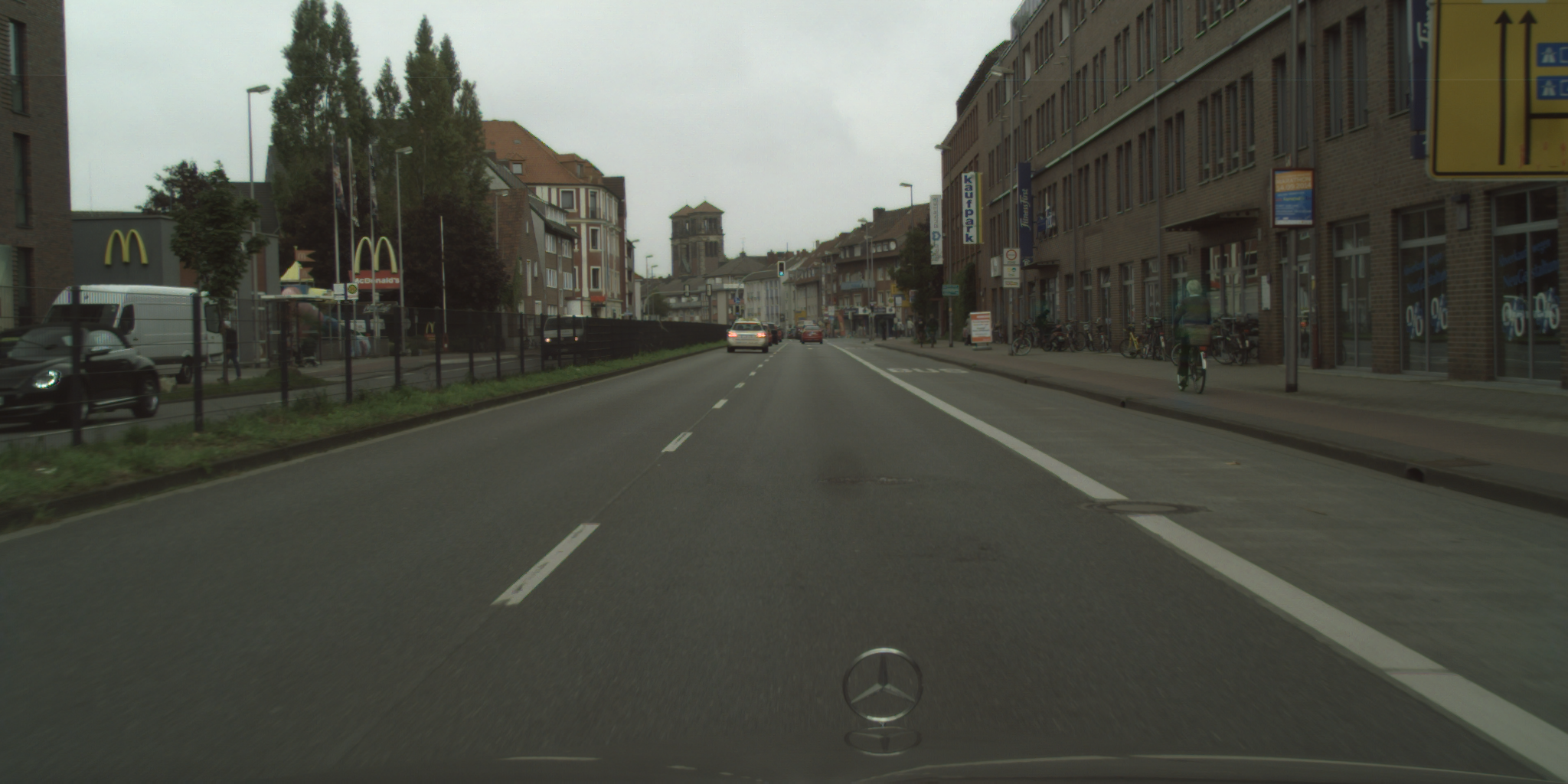}\\
    \includegraphics[width=0.3\linewidth]{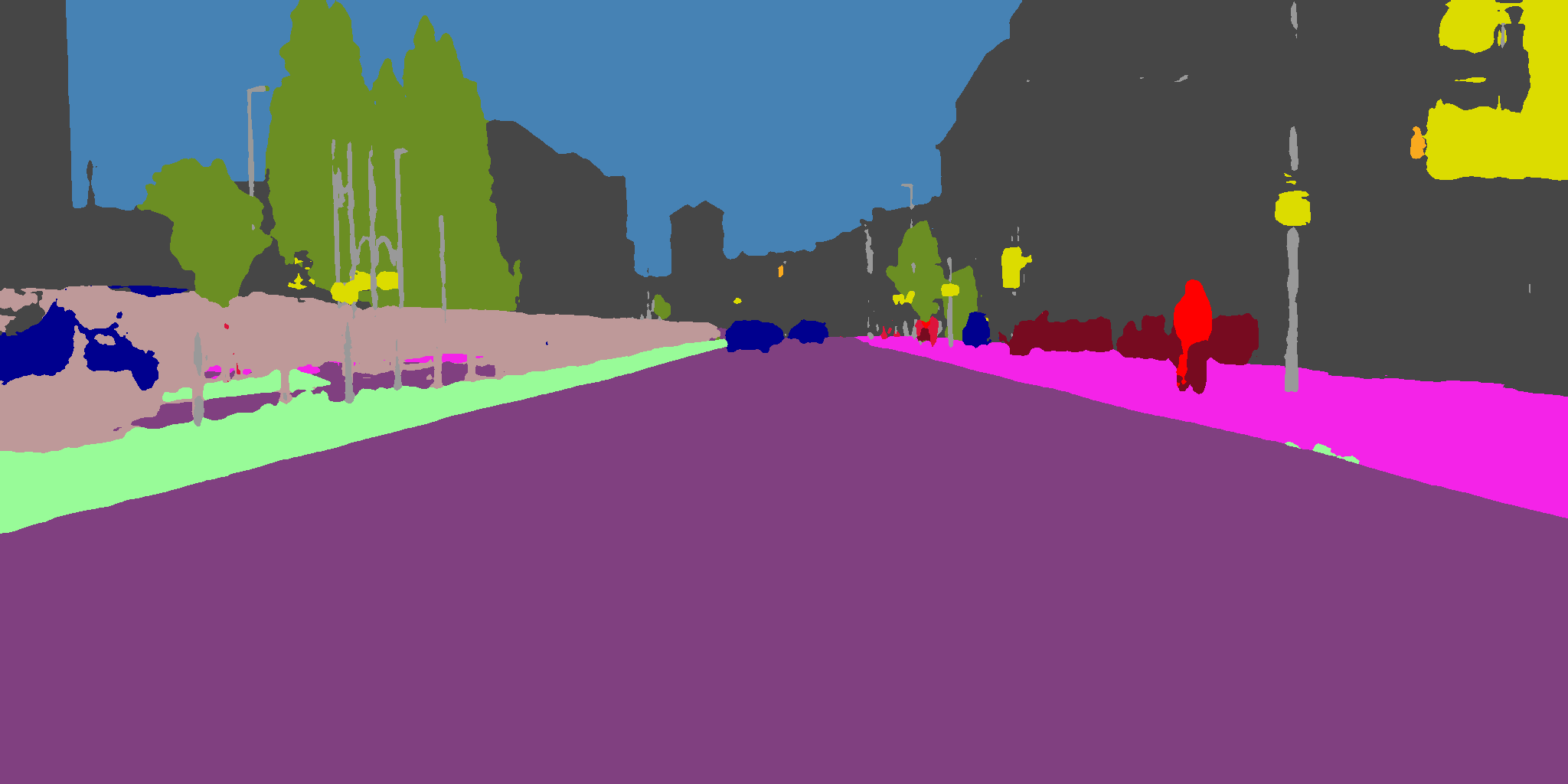}
    \includegraphics[width=0.3\linewidth]{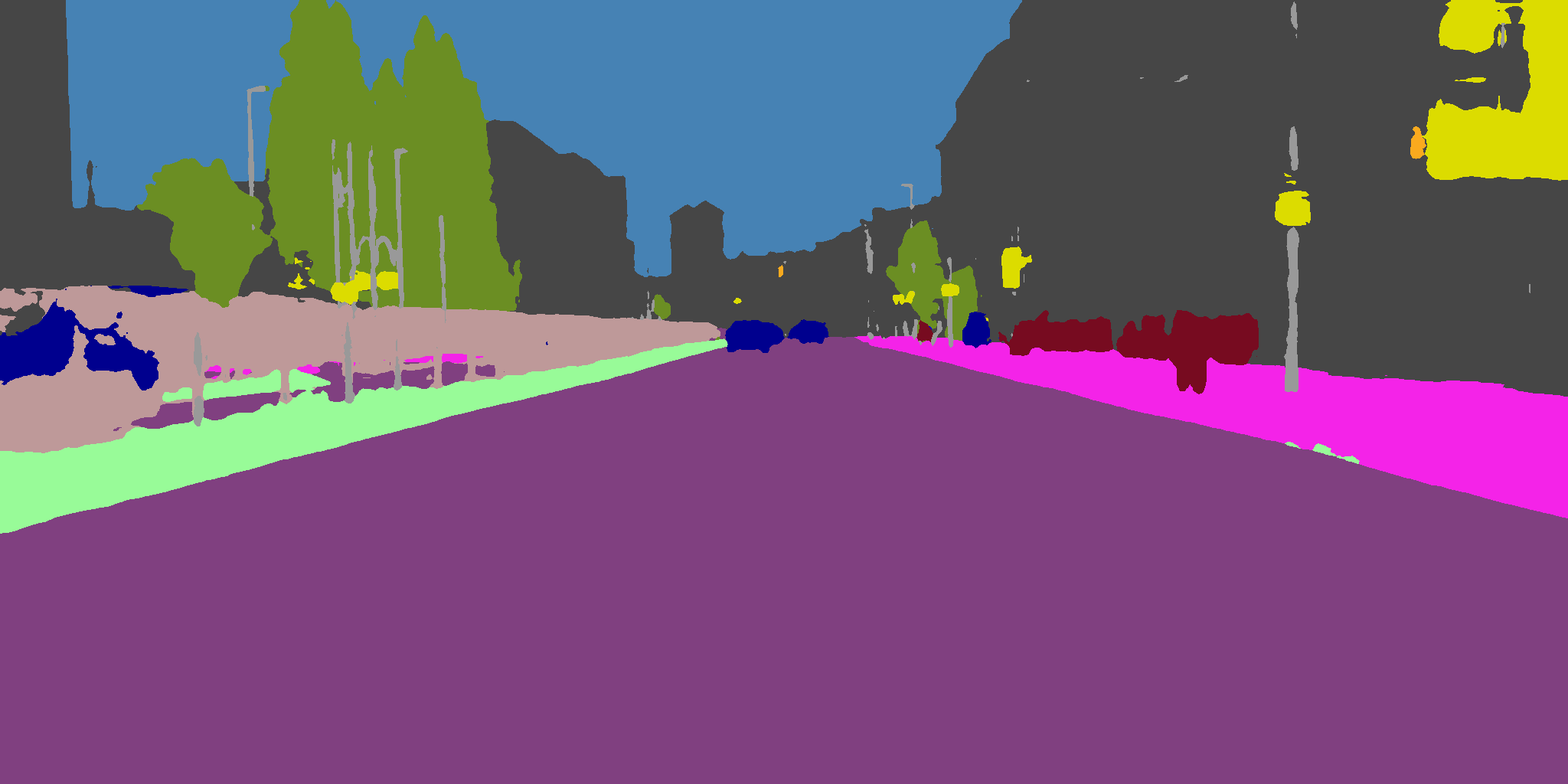}
    \includegraphics[width=0.3\linewidth]{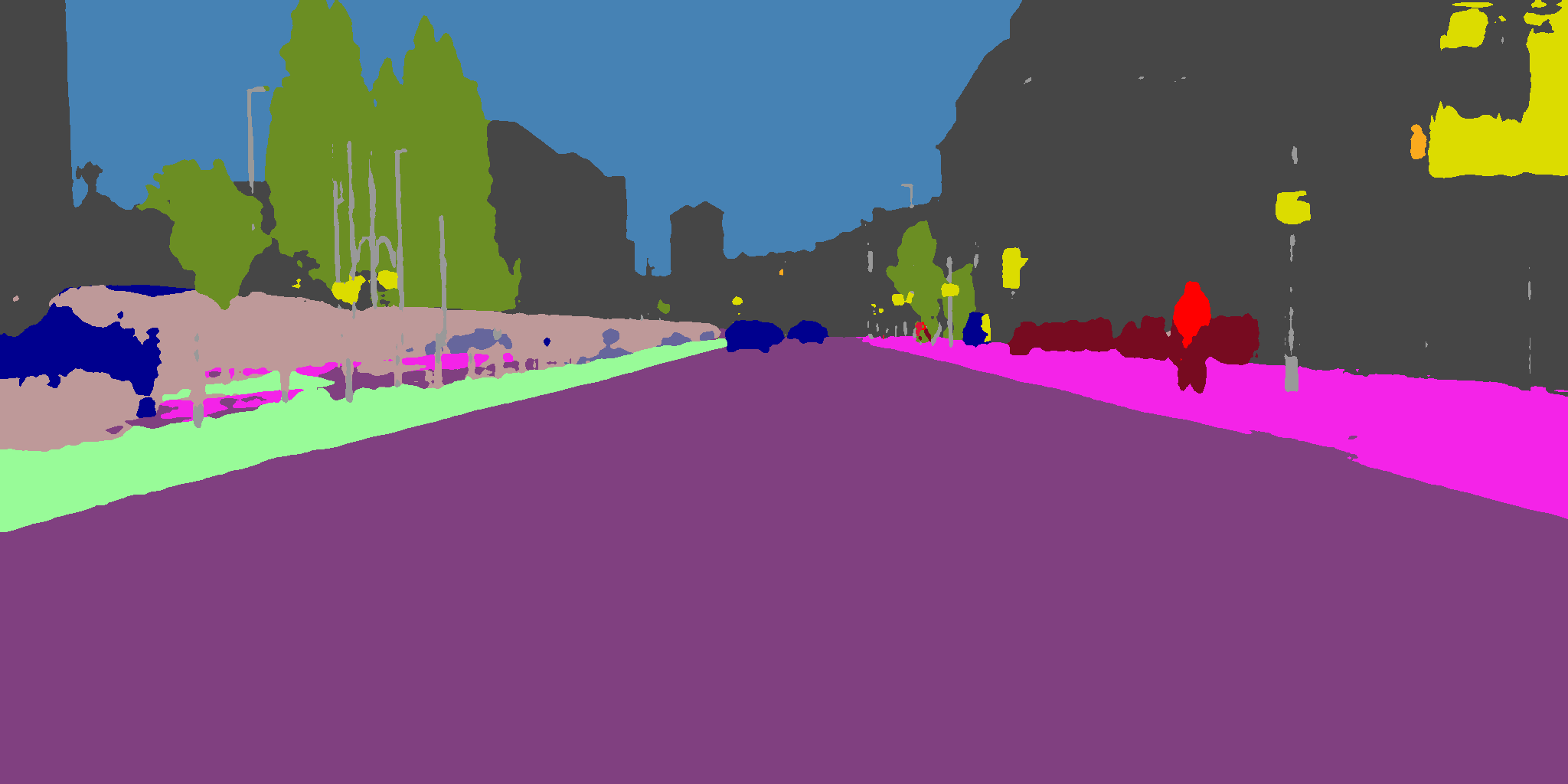}
\end{center}
   \caption{.Type\#1 attack against DeepLabV3Plus-ResNet on Cityscapes The `person' and `rider' labels vanish into the background. Our model is trained to attack DeepLabV3Plus model with MobileNet backbone, and evaluate on DeepLabV3Plus model with ResNet backbone. Top left: The input image. Top right: The input image + perturbations. Lower left: Predictions before attacks. Lower middle: Manipulated label mask. Lower right: Predictions after attacks.}
\label{fig:trained_on_mobile_test_on_resnet}
\end{figure*}

We also visualize a sample of type\#1 cross-modal attacks on Cityscape. We have two models here. The first one is trained to attack a DeepLabV3Plus-MobileNet model. The second one is trained to attack a DeepLabV3Plus-ResNet model. Then these two target models are switched. Figure~\ref{fig:trained_on_mobile_test_on_resnet} shows the first model attacks against the second model's target while Figure~\ref{fig:trained_on_resnet_test_on_mobile} shows the second model attacks against the first model's target. The results are consistent with the success rate showed in Table~\ref{gen_across_models}.

Finally, we also give the result of attacking DeepLabV3Plus-MobileNet for type\#2 and type\#3 on Cityscapes, as Figure~\ref{fig:mobile_type2_type3} shows.

\begin{figure*}[!]
\begin{center}
   \includegraphics[trim={3cm 3cm 3cm 1cm},clip,width=1\linewidth]{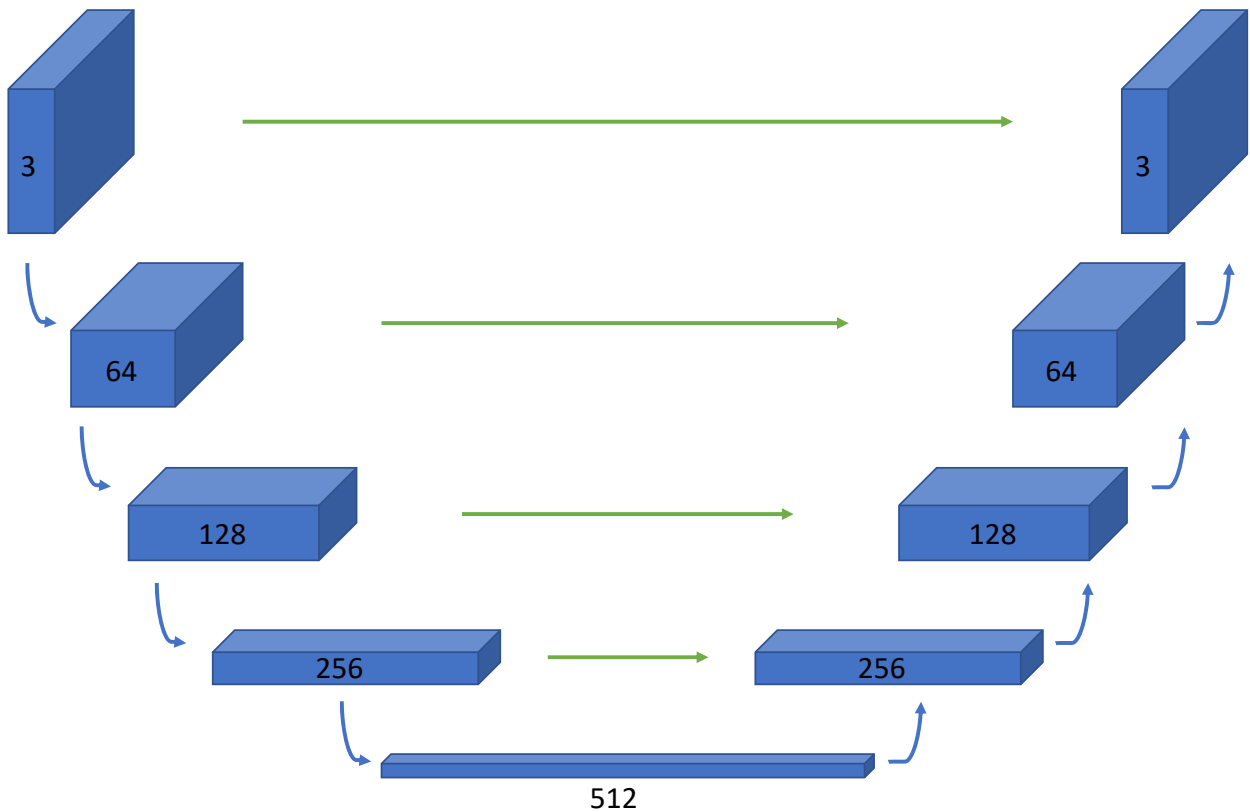}
\end{center}
   \caption{The detailed structure of the generator. 3, 64, 128, 256 are number of channels.}
\label{fig:structure}
\end{figure*}

% \begin{figure*}[t]
% \begin{center}
%     \includegraphics[width=0.3\linewidth]{images/resnet_mobilenet_10_type2/128_image.png}
%     \includegraphics[width=0.3\linewidth]{images/resnet_mobilenet_10_type2/128_pred.png}\\
%     \includegraphics[width=0.3\linewidth]{images/resnet_mobilenet_10_type3/173_image.png}
%     \includegraphics[width=0.3\linewidth]{images/resnet_mobilenet_10_type3/173_pred.png}
% \end{center}
%   \caption{CityScapes Type\#2 and Type\#3 attack on DeepLabV3Plus model with MobileNet backbone. For Type\#2 attack, the model learns to add a new set of 'person' labels in the prediction. Type\#3 attack combines both Type\#1 and Type\#2 by vanishing `person' and `rider' labels into the background, and adding a new set of 'person' labels in the prediction. Top left: The input image. Top right: The input image + perturbations. Lower left: Predictions before attacks. Lower middle: Manipulated label mask. Lower right: Predictions after attacks.}
% \label{fig:mobile_type2_type3}
% \end{figure*}